# A Theory of Consciousness from a Theoretical Computer Science Perspective: Insights from the Conscious Turing Machine

Lenore Blum and Manuel Blum[†]

## Abstract

The quest to understand consciousness, once the purview of philosophers and theologians, is now actively pursued by scientists of many stripes. We examine consciousness from the perspective of **Theoretical Computer Science (TCS)**, a branch of mathematics concerned with understanding the underlying principles of computation and complexity, including the implications and surprising consequences of resource limitations.

We are inspired by Alan Turing's simple yet powerful definition of a computer, the **Turing Machine (TM)**, and by the **Global Workspace Theory (GWT)** of consciousness originated by cognitive neuroscientist Bernard Baars and further developed by him, Stanislas Dehaene, Jean-Pierre Changeux, George Mashour and others. We are *not* looking for a complex model of the brain nor of cognition but for a simple substrate independent *computational* model of (the admittedly complex concept of) consciousness.

We do this by defining the **Conscious Turing Machine (CTM)**, also called a Conscious AI, and then we define consciousness and related notions *in the **CTM***. While these are only mathematical **TCS** definitions, we suggest *why* the **CTM** has *feelings* of consciousness. The **TCS** perspective provides a simple *framework* to employ tools from computational complexity theory and machine learning to help us understand consciousness and related concepts.

Previously we explored explanations for the feelings of pain and pleasure in the **CTM**. Here we consider additional phenomena generally associated with consciousness, again from the perspective of the **CTM**. We start with three examples related to vision (blindsight, inattentional blindness, and change blindness), then follow with a discussion of dreams, altered states, and free will. We give explanations *derived from the model* and draw confirmation from consistencies *at a high level* – well above the level of neurons – with the psychology and neuroscience literature. This paper is intended to be an introduction to a much-expanded monograph, in preparation.

---

[†] The work of Lenore Blum and Manuel Blum was supported in part by CMU, in part by a sabbatical year from CMU at the Simon's Institute for the Theory of Computing, and in part by a gift from UniDT. Email addresses: lblum@cs.cmu.edu and mblum@cs.cmu.edu . The definition of the **Conscious Turing Machine** (**CTM**) first appeared in (Blum & Blum, 2021). To be self-contained, a streamlined version of the basic definition of the model is presented here as well as amplification of key components and arguments for the "feeling of consciousness". The sections in this paper on *Blindsight, Inattentive Blindness, Change Blindness, Illusions, Dream Creation, Free Will, and Altered States of Consciousness* in the **CTM** are new. A version of this paper has been published in *PNAS* (Blum & Blum, 2022).

## Introduction: Why a Theoretical Computer Science Perspective?

Thanks to major advances in cognitive neuroscience, humanity is now on the brink of understanding how the brain achieves consciousness. In 1988, cognitive neuroscientist Bernard Baars proposed a **Global Workspace Theory (GWT)** of the brain, sketched its architecture, and outlined its implications for understanding consciousness. See (Baars B. J., 1988)[2] and (Baars B. J., 2019). That, together with the invention of fMRI in 1990, and the seminal investigations by Francis Crick and Christof Koch (Crick & Koch, 1990) into the neural correlates of consciousness, helped shake off the taboo on the scientific study of consciousness. As a consequence, the quest to understand consciousness is now actively pursued by scientists of many stripes.[3]

We study consciousness from the perspective of **Theoretical Computer Science (TCS)**, a branch of mathematics concerned with understanding the *underlying principles* of *computation*.[4] These principles largely include the *complexity* of computation, which deals with the consequences and *unexpected usefulness* of taking resource limitations into account. This perspective has provided not only a theoretical foundation for the computer revolution but also surprising new concepts and ingenious applications stemming from considerations of computational complexity. **TCS** is our principal tool. We claim that its perspective and unique insights *add* to the understanding of consciousness and related concepts such as qualia and free will. Demonstrating this is a major goal of our work.

With this in mind, we give a simple abstract substrate-independent *computational* model of consciousness that we call the **Conscious Turing Machine (CTM)** (see Chapter 1). The **CTM** is inspired by Alan Turing's simple yet powerful model of computation, the **Turing Machine (TM)**, by Bernard Baars' **GWT**, and by the **Global Neuronal Workspace Theory (GNWT)** of (Dehaene & Changeux, 2011) and (Mashour, Roelfsema, Changeux, & Dehaene, 2020).

We are not looking to model the brain or suggest neural correlates of consciousness – however interesting that may be. We are looking to understand consciousness and how a machine might experience feelings. Our intent is to come to terms, eventually, with the hard problem (Chalmers, 1995). Our view is that consciousness is a property of all properly organized computing systems, whether made of flesh and blood or metal and silicon. Confirmation for explanations from the **CTM** come from consistencies *at a high level* – well above the level of

---

[2] Baars' **GWT** is strongly influence by earlier work in cognitive science, much of which was done at Carnegie Mellon: (Simon, 1969), (Reddy, 1976), (Newell, 1990) and (Anderson, 1996).

[3] The various approaches to the study of consciousness include *psychological* (James, 1890) and (Freud S. , 1900); *philosophical* (Dennett D. C., 1991) and (Chalmers, 1996); *information theoretic* measures of consciousness (Tononi, 2004) and (Tononi & Koch, 2015); *structura*l (Baddeley & Hitch, 1974); and *neural correlates* (Dehaene & Changeux, 2011). Our approach to consciousness is *architectural*. It is informed by and close in spirit to (Baars B. J., 1997) and (Dehaene S. , Consciousness and the Brain: Deciphering How the Brain Codes Our Thoughts, 2014).

The architectural approach to the study of consciousness was inspired by the architectural models of cognition. These were developed largely at Carnegie Mellon University by Herb Simon's Sciences of the Artificial (Simon, 1969), Raj Reddy's Blackboard Model (Reddy, 1976), Allen Newell's Unified Theories of Cognition (Newell, 1990), and John Anderson's ACT-R (Anderson, 1996). The global workspace idea is due to Newell. An important more recent architectural model of cognition is LIDA (Baars & Franklin, 2009).

[4] (Sipser, 2013) is a great introduction to **TCS**.

neurons – with the psychology and neuroscience literature,[5] and from agreement with aspects of other theories of consciousness (see Chapter 5).

In this introduction, we present a brief overview of **TCS** and **CTM** using popular terms as they are typically informally understood. The perspective on **TCS** includes an example of the relevant seemingly paradoxical concept of pseudo-randomness that got defined and understood by **TCS.** In those same terms, we outline some of our understanding of consciousness from the **CTM**. These informal definitions and understandings from **CTM** are formalized after this introduction.

**The reader who wants the formal treatment straightaway can skip from here directly to Chapter 1.**

What is **Theoretical Computer Science (TCS)?** Alan Turing's seminal paper "On computable numbers" (Turing A. M., 1937) is arguably the genesis of **TCS**. That paper presents a formal mathematical definition of a "computing machine", now known as the **Turing Machine (TM)**. In it, Turing defines a simple theoretical universal programmable computer[6] that can compute any function computable by any computer or supercomputer, though of course it looks nothing like any modern-day computing machine. It was not until a decade later that Turing wrote the specs for a possible implementation.[7] Theorems are the raison d'être of mathematical theories, and Turing proved what might be called the first theorem of **TCS**, namely the *unsolvability* of the *Halting Problem*. In modern parlance, this theorem proves there can be no universal (debugger) program for determining which computer programs halt and which do not: it is just not possible to construct one. This result is equivalent to the undecidability of elementary number theory (Church, 1936), and it implies a weak form of Kurt Gödel's First Incompleteness Theorem (Gödel, 1931).[8]

After Gödel and Turing, mathematical logicians started categorizing which problems were solvable, which not, as well as investigating the esoteric hierarchy of unsolvable problems.

With the advent and wider availability of computing machines in the 1960's, it soon became clear that a number of important problems that were solvable in principle could not in fact be solved, not even with the fastest conceivable computers, and that this was not a problem with the state of technology but something deeper.[9]

---

[5] We note a historical synergy between theoretical computer science and neuroscience. Turing's simple computer model led neuroscientist Warren S. McCulloch and mathematician Walter Pitts to define their formal neuron, itself a simple model of a neuron (McCulloch & Pitts, 1943). Mathematics forced their model to have inhibition, not just excitation - because without inhibition, (loop-free) circuits of formal neurons can only compute monotonic functions - and these do not suffice to build a universal Turing Machine. The McCulloch-Pitts neuron also gave rise to the mathematical formalization of neural nets (Wikipedia, 2022) and subsequent deep learning algorithms (Goodfellow, Bengio, & Courville, 2016), further illustrating ongoing synergies.

[6] In the mathematical tradition going back to Euclid, of formulating axioms (basic premises) and proving theorems (deriving consequences), Turing identified fundamental principles (the Turing Machine) that lead to unexpected consequences (universal Turing Machine, the halting problem, etc.) and a deep understanding of computation. Many other models of discrete computation such as the lambda calculus of Alonzo Church (Church, 1936) give rise to the same set of computable functions. This gives credence to the Church-Turing thesis that no reasonable model of computation can compute more than what a Turing Machine can compute. (See (Yao, 2003) for implications to classical physics.)

[7] Although Turing's 1937 paper was strictly a paper in mathematical logic – there were no computers at the time – Turing did intend to construct a practical programmable computing machine. In 1945, he brought with him to the British National Physical Laboratory an 86-page detailed blueprint for an Automatic Computing Engine (ACE), a universal programable computer, which he intended to build (Turing A. M., 1945). Politics intervened and it was never built, only the less ambitious Pilot ACE (Hodges, 1992).

[8] Before Gödel and Turing, mathematicians had the unshakable belief that with enough knowledge and work, any mathematical problem could be solved. As the mathematician David Hilbert famously said in 1930 at his retirement address (Dawson, 1997): "This conviction of the solvability of every mathematical problem is a powerful incentive to the worker. We hear within us the perpetual call: There is the problem. Seek its solution. You can find it by pure reason, for in mathematics there is no *ignorabimus* [sic]."

[9] Suppose that when program P is run on computer C, on any input of length n, it has run time $2^n$. Let $C^+$ be the same computer except that $C^+$ runs twice as fast as C. Then program P, run on computer $C^+$, on any input of length n+1, will have run time $2^{n+1}/2 = 2^n$, which is

Researchers in the emerging field of theoretical computer science[10] realized that among natural finite (and therefore solvable) problems there appeared to be a dichotomy between those problems that were *feasibly (efficiently) solvable* and those that were not, mirroring the earlier dichotomy between solvable and unsolvable. Feasibly solvable became formalized mathematically as *solvable* (*by some computer program) in polynomial time (**P**)*. Furthermore, the realization emerged that problems solvable in polynomial time and problems *checkable in polynomial time (**NP**)* might not be equivalent.[11] Indeed, deciding the equivalence (or not) would answer the famous million-dollar ***P =? NP*** question, see (Goldreich, 2010).

Besides defining a hierarchy of serial fast (poly time) computational complexity classes, **TCS** defines a hierarchy of parallel superfast (*polylog time*) computational complexity classes. Both hierarchies inform the definitions and choices employed in our model.

Understanding the dichotomy between *easy* and *hard*, *quick* and *slow,* and their implications launched a complexity revolution with a rich theory, reframing of ideas, novel concepts and stunning applications. Indeed, developments in computational complexity over the past 40 years have shown how to use hardness to our advantage, to deal with seemingly impossible problems. This has created a paradigm shift in mathematics, namely the ability to exploit the hardness of some problems to resolve others.[12]

We illustrate with the (relevant) concept of a *computer-generated random sequence,* called a *pseudo-random sequence*.

On the face of it, the very idea of a *pseudo-random sequence* is so incongruous that von Neumann joked (von Neumann, 1951), "Anyone who considers arithmetical methods of producing random digits is, of course, in a state of sin."

More precisely, a *pseudo-random sequence generator* is a *feasible (polynomial time)* computer program for generating sequences that cannot be distinguished from truly random sequences (generated by independent tosses of a fair coin) by any feasible computer program. Thus, in the polynomial time world in which we live, pseudo-random sequences are, for all intents and purposes, truly random.[13] This understanding was impossible without the clarifications made by **TCS** and the distinctions between polynomial and superpolynomial complexity. (See (Yao, 1982) and (Yao, 2003).)

An application of the above ideas is to replace the use of random sequences in the Conscious Turing Machine (**CTM**) by sequences produced by pseudo-random generators supplied with (short) random seeds. In particular,

---

the same running time as C on any input of length n. **TCS** considers the increased speed of $C^+$ (and therefore $C^{++}$, $C^{+++}$, …) over that of C to be insignificant for running P when n is large.

[10] Early researchers in computational complexity included Jack Edmonds (Edmonds, 1965), Stephen Cook (Cook, 1971), Richard Karp (Karp, 1972), and Leonid Levin (Levin, 1973).

[11] Solvable in polynomial time (**P**) means it is possible to find a solution in time polynomial in the size of the problem instance. Checkable in polynomial time (**NP**) means that given a purported solution, its correctness can be checked in time that is polynomial in the size of the problem instance. On the face of it, finding a solution seems harder than checking it. Properly coloring the nodes of a graph with 3 colors is hard (**NP**-hard so likely not in **P**), while checking if a 3-coloring is proper, meaning that no edge joins two nodes of the same color, is easy (in **P**).

[12] Modern secure communication is based on mathematically embedding hard number theoretic problems, e.g., the integer factoring problem, in digital messages. Breaking the security would be tantamount to solving the hard problem.

[13] Such pseudo-random sequence generators are randomness amplifiers. From a short random seed, they efficiently generate long sequences that are indistinguishable in polynomial time from truly random sequences of the same length.

if the probabilistic **CTM** has "free will", as will be argued, then so does this deterministic **CTM**. This free will of a deterministic **CTM** is counter to some and perhaps much of the thinking on determinism.[14]

**Now for consciousness.** The **TCS** perspective is employed in defining the Conscious Turing Machine (**CTM**), a simple machine that formalizes a modified version of the Global Workspace Theory (**GWT**) of consciousness.

Baars describes *consciousness* through a theater analogy as the activity of actors in a play performing on the stage of Working Memory, their performance under observation by a huge audience of unconscious processors sitting in the dark (Baars B. J., 1997).[15]

In the **CTM**, the stage is represented by a Short Term Memory (**STM**) that at any moment in time contains **CTM**'s **conscious content**. The audience members are represented by an enormous collection of powerful processors – each with its own expertise – that make up **CTM's** Long Term Memory (**LTM**) (section 1.1.1). These **LTM** processors make predictions and get feedback from **CTM's** world. Based on this feedback, **learning** algorithms Internal to each processor improve that processor's behavior (section 0).

**LTM** processors, each with their own specialty, **compete** (sections 1.1.2 and 1.3) to get their questions, answers, and information in the form of **chunks** (section 1.1.3) on the stage for immediate **broadcast** to the audience. **Conscious awareness** (elsewhere called **attention**) is defined formally in the **CTM** as the reception by the **LTM** processors of the broadcast of **CTM**'s conscious content. In time, some of these processors become connected by **links** (section 1.1.4) turning **conscious communication** (through **STM**) into **unconscious communication** (through **links**) between **LTM** processors. Communication via links about a broadcasted chunk reinforces its conscious awareness, a process that Dehaene, Changeaux, et al call **ignition** (Dehaene & Changeux, 2005).

While these definitions are natural, they are merely definitions: they do not prove that the **CTM** is conscious in the sense that the term is normally used. For that, we argue that these definitions and explanations capture commonly accepted intuitive concepts of consciousness (Chapters 2) and agree, at a high-level, with cognitive neuroscience explanations of phenomena generally associated with consciousness. (Chapter 3).

Complexity considerations enter into fixing the detailed definition of **CTM**. These details include, for example,

1. the formal definition of a **chunk**, which is the information that each **LTM** processor puts into the *competition* for consciousness at each and every tick of the clock (section 1.2),
2. the fast **probabilistic competition algorithm** that selects which one of the many competing chunks reaches consciousness (sections 1.3 and 1.4),
3. the **machine learning algorithms** (section 0) in each processor that use feedback from global broadcasts, other processors, and the outside world, to update the **CTM's** competitiveness and reliability, and
4. the **memory** that in each **LTM** processor is **random access** rather than the **linear access (FILO)** of Turing machine tapes, because random access is needed for such things as fast binary search.

Complexity considerations, particularly the consequences of limited resources, play a crucial role in our high level explanations for consciousness-related phenomena such as change blindness and the feeling of free will.[16]

---

[14] "Many thinkers, indeed, believe that the determinism we find in the physical world seems to be incompatible with freedom in the sense implied by free will" (Lavazza, 2019). The fact that a random number generator can be replaced by a pseudo-random number generator equipped with a small random seed, on the other hand, suggests otherwise.

[15] See Figure 2-1 on page 42 & 43 of (Baars B. J., 1997).

[16] Eric Horvitz has pointed out the role limited resources plays for intelligent systems, see e.g., (Gershman, Horvitz, & Tenenbaum, 2015).

Although inspired by Turing's simple yet powerful model of a computer, the **CTM** is ***not*** a standard Turing Machine. That's because what gives the **CTM** its "feeling of consciousness", whether it is awake (Chapter 2) or dreaming (section 3.5), is not its computing power nor its input-output maps, but its *Global Workspace architecture,* its *predictive dynamics* (cycles of prediction, feedback and learning, section 0), its *rich multi-modal inner language* (which we call **Brainish**, section 1.2) and certain *special* **LTM** *processor*s including an **Inner generalized Speech processor** and a **Model-of-the-World processor** (Chapter 2).

As we have said, we are not looking for a model of the brain but for a simple model of consciousness, and even there, the **CTM** model can hardly be expected to explain everything: it is too simple for that. The reasonableness of the model (and its **TCS** perspective) should be judged by its contribution to the discussion and understanding of consciousness and related "hard" problems.

This paper presents an overview of the **CTM** model: we refer the reader to our first paper (Blum & Blum, 2021) for additional details. Whereas that (first) paper explores explanations for the feelings of pain and pleasure in the **CTM**[17], this (second) paper explores other phenomena generally associated with consciousness (in Chapter 3) such as dreams and free will. We also consider three examples related to vision (blindsight, inattentional blindness, and change blindness), then follow with a discussion of dreams, free will and altered states. We give explanations derived from the model and draw confirmation from consistencies at a high level with the cognitive neuroscience literature. Confirmation for the model also comes from agreement with aspects of other theories of consciousness.

These two papers are intended to be an introduction to an expanded monograph (Blum, Blum, & Blum, monograph in preparation).

[17] For an update on pain and pleasure in the **CTM**, see Chapter 4, *The Hard Problem for Pain and Pleasure*, in https://arxiv.org/abs/2011.09850.

# TABLE OF CONTENTS

# 1 The CTM Model

## 1.1 Basic CTM Structure + Definition of "Consciousness in the CTM"

Throughout this paper, statements about the Conscious Turing Machine (**CTM**) are printed in black. Examples of human and animal consciousness, generally printed in burgundy, are intended merely to clarify concepts and arguments. Burgundy-colored statements also refer to features that a human or animal *would* have if it were correctly modeled by **CTM**.

The **CTM** is (a device that is defined by) a **7**-tuple,

**< STM, LTM, Up-Tree, Down-Tree, Links, Input, Output >**,[18]

whose components are described in the rest of this section 1.1 (for more details, see (Blum & Blum, 2021)). The **CTM** has a clock that measures time in discrete clock ticks, $t = 0, 1, 2, \ldots, T \approx 10^{10}$, roughly **10** ticks per second, that being the alpha rhythm of the brain. The **CTM** has a finite lifetime, **T**. It is born at time **0**, and dies at time **T**.

### 1.1.1 Short Term Memory (STM) & Long Term Memory (LTM) Processors

In the **CTM**, the stage (conscious arena) is represented by a Short Term Memory (**STM**). This is a small memory capable of holding a single **chunk** (defined in section 1.2).

The audience (in the unconscious arena) is represented by a massive collection of $N > 10^7$ powerful parallel random-access processors**,** each with their own random-access memory, each memory large enough to hold a small multiple of **T** chunks. These together make up the **Long Term Memory** (**LTM**). Each **LTM** processor runs a number of algorithms, one of which is the processor's personal **Sleeping Experts** algorithm (section 0). All processors are in **LTM**, none in **STM**, so when we say processor, we always mean **LTM** processor.

Certain special (**LTM**) processors are particularly responsible for **CTM**'s "feeling of consciousness". These include especially a **Model-of-the-World processor** and other **Inner generalized Speech processors** for handling **inner speech, inner vision**, **inner tactile sensation**, and so on (see Chapter 2).

### 1.1.2 The Up-Tree & Down-Tree

The **Up-Tree** is an up-directed binary tree of height **h** having **N** leaves, one leaf per **LTM** processor, and a single root in **STM**. Every directed path from leaf up to the root is of length **h**. Information in the **Up-Tree** travels from the leaves below to the root above[19]. The **Down-Tree** is a simple down-directed tree of height **1** with a single root in **STM** and **N** edges directed from that root to the **N** leaves, one edge per processor, which carry information from the root to all **N** leaves.

**LTM** processors, each with their own specialty, compete via the **Up-Tree competition** (section 1.3) to get their questions, answers, and information in the form of **chunks** (section 1.2) into **STM**. The competition takes **h** clock ticks.

At each time **t**, all **LTM** processors submit information to the competition for **STM**. One of those processors wins access to **STM** at time **t+h**, and all processors receive the winning broadcast from **STM** at time **t+h+1**.

[18] Coincidently, the classical **Turing Machine** is also defined as a **7**-tuple, $< Q, \Sigma, \Gamma, \delta, q_0, q_{accept}, q_{reject} >$, where **Q** is a finite set of **States**, **Σ** is the **Input** alphabet, **Γ** is the **Tape** alphabet, δ is the **Transition** function, $q_0$ is the **Start** state, $q_{accept}$ is the **Accept** state, and $q_{reject}$ is the **Reject** state.

[19] In TCS jargon, "trees" are generally up-side down. They should perhaps have been called "roots".

### 1.1.3 Conscious Content, Conscious Awareness & Streams of Consciousness

The **chunk** that **wins** the **Up-Tree competition** (to get into **STM**) that began at time **t-h** is called *the **conscious content** of **CTM** at time **t***.[20] We say that **CTM** becomes ***consciously aware*** of that conscious content, which appeared at time **t** in **STM**, when it is received at time **t+1** by all **LTM** processors.[21]

We have defined conscious awareness as the *reception* by all **LTM** processors of **STM's** *broadcast*, rather than as the appearance in **STM** of the winning chunk*, to emphasize* that the *feeling* of consciousness arises after all processors, including especially the **Model-of-the-World** and **Inner generalized Speech processors** (see Chapter 2) receive the broadcast and act on it.

Our *definition* of **conscious awareness** in the **CTM** (i.e., the *reception* by all **LTM** processors of **STM's** *broadcast*) aligns roughly with what cognitive neuroscientists call "attention". See for example (Graziano, Guterstam, Bio, & Wilterson, 2020) and (Mashour, Roelfsema, Changeux, & Dehaene, 2020). Reverberation of chunks via processor links immediately following reception is related to what (Dehaene & Changeux, 2005) call "ignition". What cognitive neuroscientists call "awareness", or "subjective consciousness", align roughly with what we call "the *feeling* of consciousness" in the **CTM** (see Chapter 2).

One reason to keep the number of chunks in **STM** small (exactly one in our model) is to ensure that all processors focus on the same information in the broadcast from **STM**. Another reason is to keep the model as simple as is reasonably possible for an understanding of consciousness.[22]

**CTM** is constantly bubbling with the activity of chunks competing for **STM**, its winners being (constantly) broadcast from **STM** to **LTM.**[23] The time-ordered chunks that are broadcast from **STM** to **LTM** form **a *stream of consciousness***. This stream, as argued in Chapter 2, is part of the subjective "feeling of consciousness".

### 1.1.4 Links, Unconscious Communication & Global Ignition

All communication between any two processors occurs *initially* (at birth and until links are formed) via **STM**. For example, processor **A** can submit a query to the **Up-Tree competition**. If the query wins the competition, it is broadcast to all processors. Processor **B** may then submit an answer via the competition, which if it wins gets

---

[20] For simplicity, **STM** holds only one chunk at any moment in time. In humans, the storage capacity of short-term memory is roughly 7±2 chunks (Miller, 1956), where a chunk can be a word, a phrase, a digit, and so on. A few chunks cycling through **STM** can simulate some aspects of an **STM** that holds several chunks. Cycling can happen via the **Up-Tree competition** and the **Down-Tree broadcasts. I**n this way, **CTM** can keep thoughts alive in **STM** continuously through many cycles by sending the thought from **processor** to **STM** to **processors** to **STM** to ….

[21] For simplicity, we stipulate reception of the broadcast by *all* **LTM** processors. This is a simplification of what goes on in humans, as the dorsal stream of vision is never conscious, only the ventral stream is conscious.

[22] Leslie Valiant (Valiant, 2013, pp. 127-128) does not assert that focus is primary. He asserts instead that limited computational resources and constraints imposed by the need to learn are the primary reasons for the small size of conscious information. While these are reasonable factors, the principle factor in our opinion is the required focus.

[23] This bottom-up/top-down cycle is analogous to the Global Neuronal Workspace (**GNW**) hypothesis (Dehaene, Changeux, & Naccache, 2011) that "conscious access proceeds in two successive phases … . In a first phase, lasting from ≈100 to ≈300 ms, the stimulus climbs up the cortical hierarchy of processors in a primarily bottom–up and non-conscious manner. In a second phase, if the stimulus is selected for its adequacy to current goals and attention state, it is amplified in a top–down manner and becomes maintained by sustained activity of a fraction of **GNW** neurons, the rest being inhibited. The entire workspace is globally interconnected in such a way that only one such conscious representation can be active at any given time … ."

The d*ynamic* of chunk submission to competition for **STM** (the stage) and subsequent broadcast of the winning chunk to the **LTM** processors (the audience) corresponds roughly to the **GNW** global *ignition* property. However, **GNW** ignition is significantly more subtle and complicated, depending, e.g., on strength (or absence) of sensory inputs), and having variable duration.

broadcast, and so on. If **A** acknowledges that **B**'s answer is useful, and this occurs sufficiently often, then a bi-directional **link** forms (or an existing one is strengthened) between **A** and **B**.[24]

In addition to processors putting chunks into the **Up-Tree competition**, processors can send chunks through **links**. When chunks are sent through **links** that formerly went through **STM**, ***conscious communication*** (through **STM**) between **A** and **B** turns into ***unconscious communication*** between **A** and **B**.[25] As additional links form between **A** and **B** we say the link between **A** and **B** is ***strengthened***.

**Links** are channels for transmitting information between processors. Those chunks sent between linked processors following the broadcast of **CTM**'s **conscious content**, can re-enforce and sustain **conscious awareness**.[26] This re-enforcement is related to what (Dehaene & Changeux, 2005) call "global ignition" in their Global Neuronal Workspace Theory (**GNWT**).

### 1.1.5 Input & Output Maps, Sensors & Actuators

**CTM**'s **environment (Env)** is a subset of $\mathbf{R^m(t)}$ where **R** denotes the real numbers, **m** is a positive integer dimension, and **t**, a non-negative integer, is time.

**Input** maps take (time-varying) environmental information acquired by **CTM**'s **sensors**, and send it to designated processors to convert into chunks (section 1.2).

**Output** maps take chunks from designated processors, convert them to command instructions, and send those instructions to **actuators** that act on the **environment**.

### 1.1.6 Summary of Connections

In summary, there are five kinds of connections in the **CTM** that provide paths and mechanisms for transmitting information. The five, shown in *Figure 1*, are:

1. **Env → LTM:** directed edges from the environment via **sensors** to processors of the sensory data.
2. **LTM → STM:** via the **Up-Tree** edges.
3. **STM → LTM:** via the **Down-Tree** edges.
4. **LTM → LTM:** bi-directional edges (**links**) between processors.
5. **LTM → Env:** directed edges from specific processors (like those that generate instructions for finger movement) to the environment via **actuators** (like the fingers that receive instructions from these processors) that act on the environment.

---

[24] Linking is reminiscent of the Hebbian principle (Hebb, 1949) "Neurons that fire together wire together".

[25] If the **CTM** has $10^7$ **LTM** processors, corresponding to ≈$10^7$ cortical columns in the brain, and if all pairs of processors are initially linked, there will be ≈$10^{14}$ links, which is a large possibly infeasible number of links: in the brain, a link is an axon, part of a neuron, at most one axon per neuron, and since the brain has less than $10^{11}$ neurons, it has less than $10^{11}$ links, so not all processors can be linked.

Worse yet, a processor with $10^7$ inputs might need to run its own personal competition to decide what to look at.

Although one might want certain special processors linked initially, we choose for simplicity to have no processors linked at birth.

[25] The question arises: Why is a tree necessary? Why not just compute $\mathbf{f(chunk_{p,t,0}) / \sum_{all\ N\ LTM\ processors\ p'} f(chunk_{p',t,0})}$ in.

[26] "Global ignition occurs when a broadcast excitation exceeds a threshold and becomes self-reinforcing, with some neurons exciting others that, in turn, return the excitation. The connected burst into a self-sustained state of high level activity, a reverberating 'cell assembly,' as Hebb called it," is what (Dehaene S. , Consciousness and the Brain: Deciphering How the Brain Codes Our Thoughts, 2014) calls global ignition.

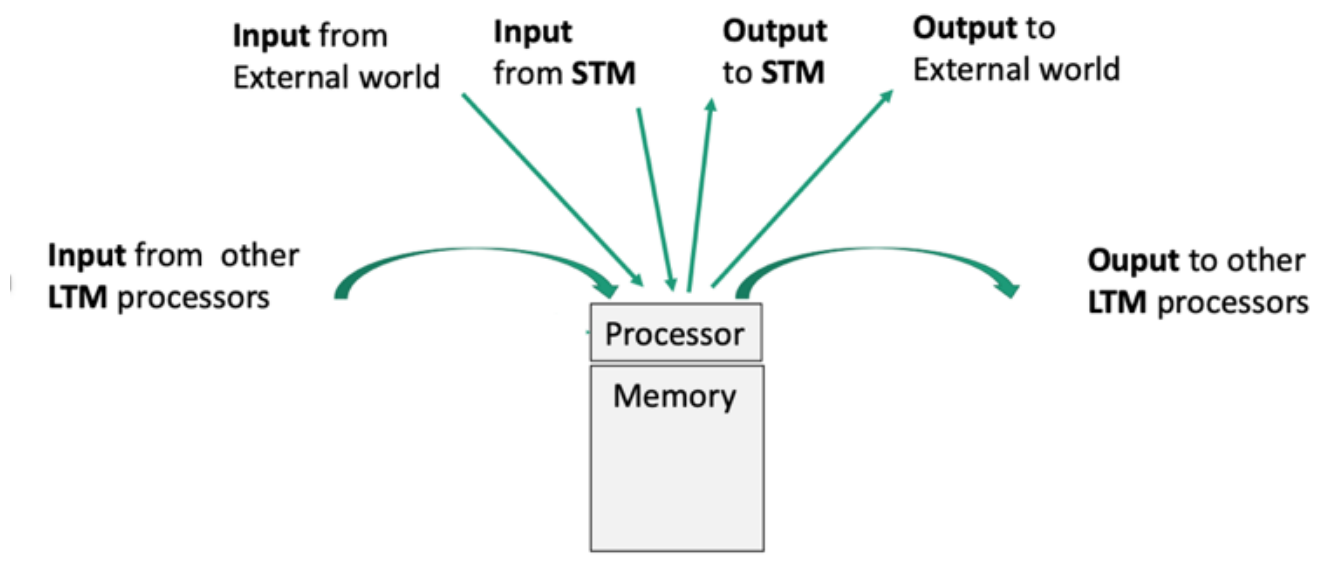

*Figure 1 Edge connections in **CTM** to and from an **LTM processor**.*

## 1.2 Brainish (the CTM's Multi-Modal Inner Language), Chunks & Gists

**Brainish** is **CTM**'s **inner language** used (by the gists in chunks) to communicate *between* **processors**,[27] whether via the competition and broadcasts or directly through **links**. Brainish is the language used to express **inner speech**, **inner vision**, **inner sensations** (Chapter 2), **imaginings** and **dreams** (section3.5 ). It includes coded representations of **inputs** and **outputs** all expressed with succinct **multi-modal** Brainish words and phrases called **gists**. A gist can hold the essence of a scene or the (high level expandable) idea of a proof. It can be an answer to a query, an insight of some sort, a dream image, a description of pain, and so on. Brainish is able to express and manipulate images, sounds, tactile sensations, and thoughts – including unsymbolized thoughts – *better* than **spoken outer languages** such as English, Chinese or "doggish" (Slobodchikoff, 2012). We claim that having an expressive inner language is an important component of the *feeling* of consciousness (Chapter 2).[28]

Information is carried on all edges between **LTM** processors, between **STM** and **LTM**, from **input** to **LTM**, and from **LTM** to **output**, by **chunks**.

A **chunk** is a **6**-tuple, **< address, t, gist, weight, intensity, mood >**, where

- **address** is the address[29] of the **LTM** processor that produced the chunk,
- **t** is the time that the chunk got produced (put into the competition),
- **gist** is the information, "concisely expressed" in **Brainish**, that the processor intends to communicate. A **gist** can hold the essence of a scene or the (high level expandable) idea of a proof. It can be an answer to a query, an insight of some sort, a dream image, a description of pain such as from a torn ligament, and so on.
- **weight** is a valanced real number that the processor gives the gist,
- **intensity** starts off as **|weight|**, and
- **mood** starts off as **weight**.

We note that the **size** of a chunk (and hence the size of its components, including its gist) will necessarily be bounded by computational complexity considerations to be described (see section 1.4 and (Blum & Blum, 2021) for more specifics).

---

[27] The language used internally by a processor, as opposed to between processors, varies in general from one processor to another; it includes but is not restricted to Brainish.

[28] Representing and learning multimodal inputs is an important aspect of machine learning **(ML**). Paul Liang, PhD student in ML at Carnegie Mellon, is developing Brainish, **CTM**'s multimodal language.

[29] If the **CTM** has $10^7$ **LTM** processors, corresponding to ≈$10^7$ cortical columns in the brain, then the address is a 7 digit number.

## 1.3 (Probabilistic) Up-Tree Competition, Coin-Flip Neuron & Competition Functions

The **Up-Tree competition** is the mechanism that determines which **LTM** processor will get its **chunk** into **STM**. At each clock tick **t = 0, 1, …., T**, the $t^{th}$ competition starts with each processor **p** putting its chunk, with its time set to **t**, into its leaf of the **Up-Tree**. After a chunk is submitted to the **Up-Tree** competition and while it moves up the competition tree, its **address, t, gist,** and **weight** remain unchanged; but its **intensity** and **mood** get updated to incorporate ever more global information.

Deciding whether or not a chunk (a variant of the original chunk) moves up one level in the **Up-Tree** or drops out is made by a fast tiny parallel circuit, one such circuit located in each of the non-leaf nodes of the **Up-Tree**, each making its decision in one clock tick (the time between two successive clock ticks).

The **probabilistic Up-Tree**, rather than the **deterministic Up-Tree**, is the right one to consider[30]. In the **probabilistic Up-Tree**, each node has and uses a **coin-flip neuron** (*Figure 2*) in its built-in circuit. A **coin-flip neuron** is a device that takes as input an (ordered) pair **(a, b)** of non-negative real numbers (**a ≥ 0** and **b ≥ 0**), and *in one step* – a fraction of a clock tick – does the following:

if **a+b > 0**, it outputs **a** with probability **a/(a+b)**, else **b**; ; if **a = b = 0**, it outputs **a** with probability **½**, else **b**.

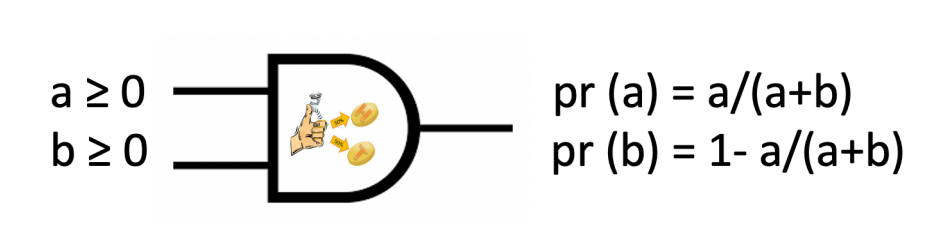

*Figure 2 A coin-flip neuron on an input **(a, b)** with **a + b > 0.***

At each clock tick, the circuit in a non-leaf node **v** runs a *local* competition that probabilistically selects one of **v**'s two (sibling) children based on a comparison of the chunks they contain, then moves (a variant of) that chunk into **v**. That chunk is said to be the **winner of the local competition** at/for **v**.

The local competition employs the **Up-Tree's** (i.e., **CTM**'s) **competition function f**, a function that maps chunks to non-negative real numbers in a fraction of a clock tick, to choose the local winner.

Specifically, suppose at level **s**, **0 < s ≤ h**, node $v_s$ at level **s**, $chunk_{p(L)}$ and $chunk_{p(R)}$ are the chunks in $v_s$'s left and right children respectively. Then:

1. with probability **{ f(chunk** $_{p(L)}$**) / (f(chunk** $_{p(L)}$**) + f(chunk** $_{p(R)}$**))** if the denominator ≠ **0**, or with probability ½ if the denominator = **0 }**, $chunk_{p(L)}$ is the **local winner**,
2. else $chunk_{p(R)}$ is the local winner.

We now specify the chunk that moves into $v_s$. Suppose the local winner at $v_s$ is the variant of $chunk_{p,t,0}$ having $address_p$, **t**, $gist_{p,t,0}$, and $weight_{p,t,0}$. Then the chunk that moves into $v_s$ will be:

$$chunk_{p,t,s} = < address_p, t, gist_{p,t,s}, weight_{p,t,s}, intensity_{p,t,s}, mood_{p,t,s} >, \text{ where}$$

$gist_{p,t,s} = gist_{p,t,0}$, $weight_{p,t,s} = weight_{p,t,0}$, $intensity_{p,t,s} = (intensity_{p(L),t,s-1}) + (intensity_{p(R),t,s-1})$, and $mood_{p,t,s} = (mood_{p(L),t,s-1}) + (mood_{p(R),t,s-1})$.

Note that $intensity_{p,t,s} = \sum_{p'} (intensity_{p',t,0})$ and $mood_{p,t,s} = \sum_{p'} (mood_{p',t,0})$, where the two sums $\sum_{p'}$ are over all **LTM** processors **p'** in the subtree rooted at $v_s$.

[30] For a partial explanation, see **Q7** of the **FAQ**.

We call a **CTM** with a probabilistic **Up-Tree** competition a ***probabilistic CTM***. In this paper, except where otherwise stated, all **CTM**s will from now on be probabilistic.

**REMARK.** Updating the chunk at node $v_s$ consists in computing the **f**-values of the two sibling children of $v_s$, using that as input to a **coin-flip neuron** to choose the local winner, and then making the needed additions to get **intensity** and **mood**. This must all be done in **1 clock tick**, which puts bounds on both the *amount of computation* that can be performed in a node and *the size* of the **chunk** in that node. (See section 1.4 and (Blum & Blum, 2021) for more specifics.)

By a simple induction, the **winner of the Up-Tree competition** (the *conscious content* of **CTM** at time **t+h**) will be:

$$\mathbf{chunk}_{p,t,h} = <\mathbf{address}_p, \mathbf{t}, \mathbf{gist}_{p,t,0}, \mathbf{weight}_{p,t,0}, \mathbf{intensity}_{p,t,h}, \mathbf{mood}_{p,t,h}>$$

where $\mathbf{intensity}_{p,t,h} = \sum_{\text{all N processors p' in LTM}} (\mathbf{intensity}_{p',t,0})$ and $\mathbf{mood}_{p,t,s} = \sum_{\text{all N processors p' in LTM}} (\mathbf{mood}_{p',t,0})$.[31]

Let **t ≥ h**. The ***current mood*** of **CTM** at time **t**, $\boldsymbol{mood_t}$, is defined to be the **mood** of the **chunk** that is broadcast from **STM** at time **t**. Thus **CTM** becomes *consciously aware* of $\mathbf{mood}_t$ at time **t + 1**.

**REMARK.** $\mathbf{mood}_t = \sum_{\text{all N LTM processors p}} \mathbf{mood}_{p,t-h,0}$, so $\mathbf{mood}_t/\mathbf{N}$ is the average mood of the chunks submitted to the competition at time **t - h**.

$\boldsymbol{Mood_t}$ is defined to be **CTM**'s mood, "optimistic/happy" if positive, "pessimistic/sad" if negative, at time **t** when the winning chunk is in **STM.**[32] (Alternatively, $\mathbf{Mood}_t$ could have been defined to be the mood at time **t-h** when the winning chunk was put into the competition, or the mood at time **t+1** when the chunk was received by all processors by broadcast). $\boldsymbol{Intensity_t}$ is defined similarly to be **CTM**'s level of "energy/enthusiasm/confidence" at time **t**.

We say that a competition function **f** is ***additive*** if $\mathbf{f(chunk}_{p,t,s}\mathbf{(v_s)) = f(chunk_L(v_s)) + f(chunk_R(v_s))}$. Examples of additive competition functions include $\mathbf{f(chunk}_{p,t,s}) = \mathbf{intensity}_{p,t,s}$, or more generally,

$\mathbf{f(chunk}_{p,t,s}) = \mathbf{intensity}_{p,t,s} + \mathbf{c}\cdot\mathbf{mood}_{p,t,s}$ for any real **c**, $-1 \le c \le +1$, but not $\mathbf{f(chunk}_{p,t,s}) = |\mathbf{mood}_{p,t,s}|$.

$\mathbf{f(chunk}_{p,t,s}) = |\mathbf{mood}_{p,t,s}|$ is ***not*** additive, because $|a + -a| \neq |a| + |-a|$.

**THEOREM.** If the **competition function f** of a **probabilistic CTM** is **additive**, then every chunk submitted to the **Up-Tree competition** gets a fraction of time in **STM** proportional to its **f-value**, its importance as determined by **f**. Specifically, the probability that a submitted $\mathbf{chunk}_{p,t,0}$ gets into **STM** is

$$\mathbf{f(chunk}_{p,t,0})/\textstyle\sum_{\text{all N LTM processors p'}} \mathbf{f(chunk}_{p',t,0}).$$ [33]

As a consequence, for additive **f**, the permutation chosen to assign processors to leaves of the **Up-Tree** has no effect on the sequence of broadcasts from **STM** (see (Blum & Blum, 2021) for more specifics including the statements of these theorems and their proofs).

---

[31] The question arises: Why not just compute $\mathbf{intensity}_{p,t,h} = \sum_{\text{all LTM processors p'}} (\mathbf{intensity}_{p',t,0})$ in one step to get the intensity of the chunk winning the competition? (Similar question for $\mathbf{mood}_{p,t,s} = \sum_{\text{all LTM processors p'}} (\mathbf{mood}_{p',t,0})$.) Answer: however you do it, you need log **N** steps to compute the ∑. Note that processors in general do not know what chunks other processors have entered into the competition. Global information is accumulated locally at each step of the **Up-Tree competition**.

[32] (Kringelbach & Berridge, 2017) argue that in humans "emotion is always valenced—either pleasant or unpleasant—and dependent on the pleasure system".

[33] The question again arises: Why not just compute $\mathbf{f(chunk}_{p,t,0}) / \sum_{\text{all N LTM processors p'}} \mathbf{f(chunk}_{p',t,0})$ in one step to get the probability of a chunk winning the competition? One of many answers: however you do it, you need $\mathbf{log_2 N}$ steps to compute the ∑.

We note that **f(chunk$_{p,t,s}$) = |mood$_{p,t,s}$|** is a bad choice for more reasons than that it is not additive: If at level **0,** two sibling chunks have **weight**s **+100** and **-100** and all other chunks have **weight** = +1, then neither of the two high **intensity** siblings will reach **STM**. This might occur if you spy a $10 bill on the ground, but someone else picks it up. In that case, the pleasure of finding the $10 bill and the pain of losing it will never reach consciousness. You would be unconscious of both. This seems unlikely to us.

**EXAMPLE.** let **f = |mood|**. Set **$w_1$ = 100, $w_2$ = -100, $w_3$ = 1, $w_4$ = 2**. Then the **competition Up-Tree** looks like this:

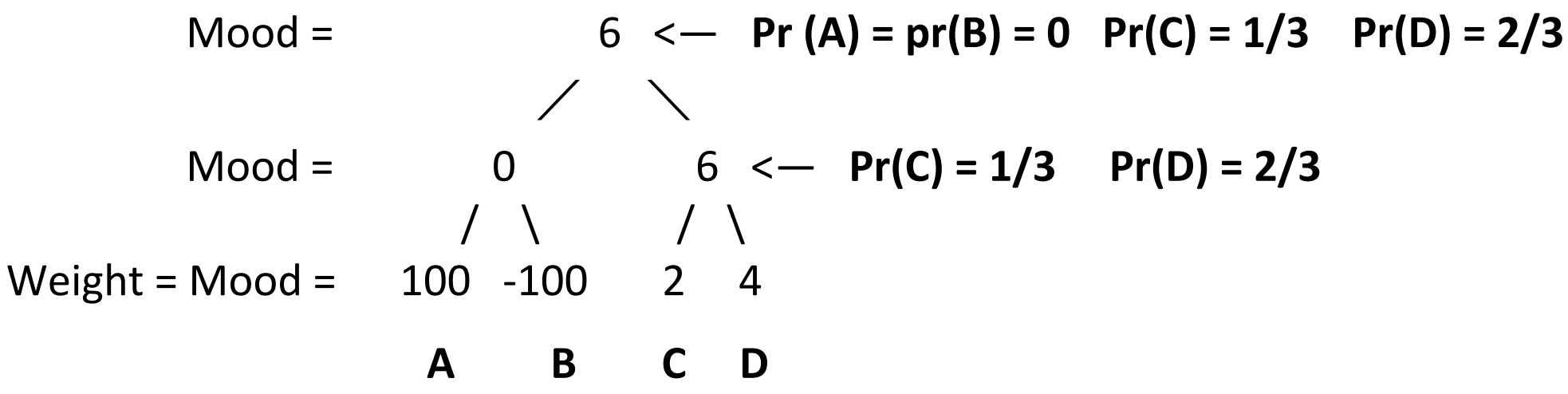

*Figure 3 Up-Tree competition for f(chunk) = |mood|.*

## 1.4 Complexity of Computation & Time Delay for Conscious Awareness

For **t ≥ 0** and **s > 0**, the *computation* to update the chunk at node **$v_s$** in the **Up-Tree competition** consists of:

1. two **fast** computations of **f**, a sum and division of their values, and a fast probabilistic selection,[34]
2. putting the **address, gist** and **weight** of the selected chunk into **$v_s$**, and
3. summing the **intensities** and **moods** of the chunks associated with **$v_s$**'s children, and setting those sums to be the **intensity** and **mood** respectively of the chunk at **$v_s$**

These computations, all three of which must be completed in **1 clock tick** put a bound on both *the size* of the **chunk** in a node and the *amount of computation* that can be performed in that node.[35]

If **1 clock tick** is **100ms** long and there are **$10^7$ LTM** processors, then the time from a chunk being placed into competition by an (unconscious) **LTM** processor to becoming **CTM's** *conscious content* will be about 2.3 seconds. If the broadcast takes another 100ms, the total time to *conscious awareness* will be about 2.4 second. Curiously, this seems to be about how much time it takes for humans to become consciously aware of decisions made unconsciously (Smith, 2008). This time can be reduced from 2.4 seconds to .7 sec if the interior nodes of the up-tree each have 10 children instead of 2. (See section 3.7 for implications of an **Up-Tree** delay on free will in the **CTM** and/or humans.)

## 1.5 Memories & The High Level Story

We assume that each processor **p** stores in its internal memory at time **t**, the **chunk$_{p,t,0}$** that it submitted to the competition, the chunk it received by broadcast from **STM**, and a select subset[36] of chunks it received from **links** or from **Input maps**. These stores are a substantial part of **CTM**'s ***memories***.

---

[34] Alternatively, if **f** is included with the chunk in the node, then a sum and a fast probabilistic selection will do it faster.

[35] The space required to store a chunk must be large enough to store a **$\log_2 N$** bit address, and to store a **gist** whose length is no greater than what is required to store approximately one line of English or its equivalent in Brainish, very roughly $2^{10}$ bits.

[36] Assuming **CTM** is a good model for the brain, it is not possible for each of its **N ≈ $10^7$** processors to record from all other **N-1** processors, as that would require **CTM** to have more links than there are neurons in the brain.

This "history" provides a **high level story** of what **p** saw and did. High level stories account in large part for **CTM**'s sense of self in its "feeling of consciousness" (Chapter 2). They are called upon when **CTM** creates dreams (section 3.5).

Periodically, this stored information may be pruned so only "salient" chunks remain, the most "salient" being those that represent terrible, wonderful, or unexpected events.[37] In general (see section 0), every processor makes predictions regarding the chunks it generates, modifies, reconstructs, reconsolidates, and stores.

## 1.6 Predictive Dynamics = Prediction + Feedback + Learning (Sleeping Experts Algorithm)

Processors require **feedback** to *assess correctness, detect errors*, and **learn** how to boost correctness (diminish and *correct* errors).

- **Predictions** in **CTM** are made by **LTM** processors both within their internal algorithms and implicitly when they submit chunks elsewhere, whether to the competition for **STM**, to other processors through **links**, or to **actuators** that effect the **environment**.
- **Feedback** comes from chunks that are received in broadcasts from **STM**, through **links**, and from sensors of the **environment** via **Input maps**, indicating correctness or detecting errors in predictions.
- **All learning** and **error correcting** take place within processors.

There is a continuous cycling of prediction, feedback and learning within **CTM**. The **CTM** needs to be alert to anything unusual, surprises of any kind, in order to deal with such things if necessary and to improve its understanding of the world in any case. Prediction errors (e.g., "surprises") are minimized by this cycling.

Processors especially need to know if they were too timid or too bold in setting their **|weights|** so they can correct their weight-assigning algorithms. **Sleeping Experts Algorithms** are a class of **learning algorithms** employed by **LTM** processors to do just that. See (Blum A. , 1995) and (Blum, Hopcroft, & Kannan, 2015) for one of the simplest versions of the **Sleeping Experts Algorithms (SEA)**.[38] Here is the idea:

In general, **SEA** will *embolden* its processor – pushing it to raise the *intensity* it gives its chunks – if

1. its chunk did not get into **STM**, *and*
2. its information is more valuable (in the **SEA**'s opinion) than what got into **STM**.

The **SEA** will *hush* its processor – pushing it to lower the *intensity* it gives its chunks – if

1. its chunk got into **STM**, and
2. the information in that chunk is found to be less valuable than that of another chunk that failed to get into **STM**. (This hushing will typically occur sometime later when the **SEA** becomes aware of the more highly valued information.)

**Sleeping Experts Algorithms** play a role in whether or not processors get their chunks into **STM**. They also play a role in whether or not processors "pay attention" to gists in chunks that are sent to them via **links**. The **|weight|** of a chunk is an indication of how important the processor that created the chunk believes its gist to be, and this will influence whether or not a processor that receives the chunk will pay attention to it.

---

[37] Might the **CTM** also use compression for this storage? (Al Roumi, Marti, Wang, Amalric, & Dehaene, 2020) discuss mental compression of spatial sequences in human memory.

[38] More sophisticated **Sleeping Experts Algorithms** will be presented in an expanded monograph (Blum, Blum, & Blum, monograph in preparation). See also, (Blum A. , 1995), (Blum A. , 1997), (Freund, Schapire, Singer, & Warmuth, 1999), (Blum & Mansour, 2007), (Luo & Schapire, 2015) and (Blum, Hopcroft, & Kannan, 2015).

## 1.7 Comparison of CTM with the Global Workspace Theory Model

We conclude this chapter with a comparison between the **CTM** and Baars' **GWT** model (*Figure 4*).

Aiming for simplicity, we have eliminated or simplified many features in **GWT.** For example, the **CTM** has just one "actor" on stage holding just one chunk at a time. Additionally, in the **CTM**, all processors are in **LTM**. We have eliminated the Central Executive since its sequence of directions, opinions, questions, answers and so on, as all this can come from processors. In the **CTM, inputs** and **outputs** go directly to and from **LTM** processors, not directly through **STM**. This modification is consistent with the neuroscience literature (Liljenström, 2021).

In the **CTM, chunks** compete in a *well-defined* **competition** to get onto the stage (**STM**). *Conscious awareness (attention)* is the reception by all **LTM** processors of the broadcasted **winning chunk** (i.e., **CTM**'s *conscious content*), not an event that occurs between **Input** and **STM**. The roles of Baddeley and Hitch's Verbal Rehearsal and Visuospatial Sketchpads (Baddeley & Hitch, 1974) are assumed by **LTM** processors.

*Predictive dynamics* (cycles of prediction, feedback and learning), a multi-modal inner language (*Brainish*) as well as computational and complexity considerations, are explicit key **CTM** features.

Finally, as in the "Extended Mind Theory" of (Clark & Chalmers, 1998), **CTM** can have access to existing technology such as Google, Wikipedia, WolframAlpha, AlphaGo, and so on, in the form of **LTM** processors tasked to use these apps. This is one way to ensure that **CTM** has a huge collection of powerful processors at the start of its life (**t = 0**), a collection that is augmentable throughout its life.

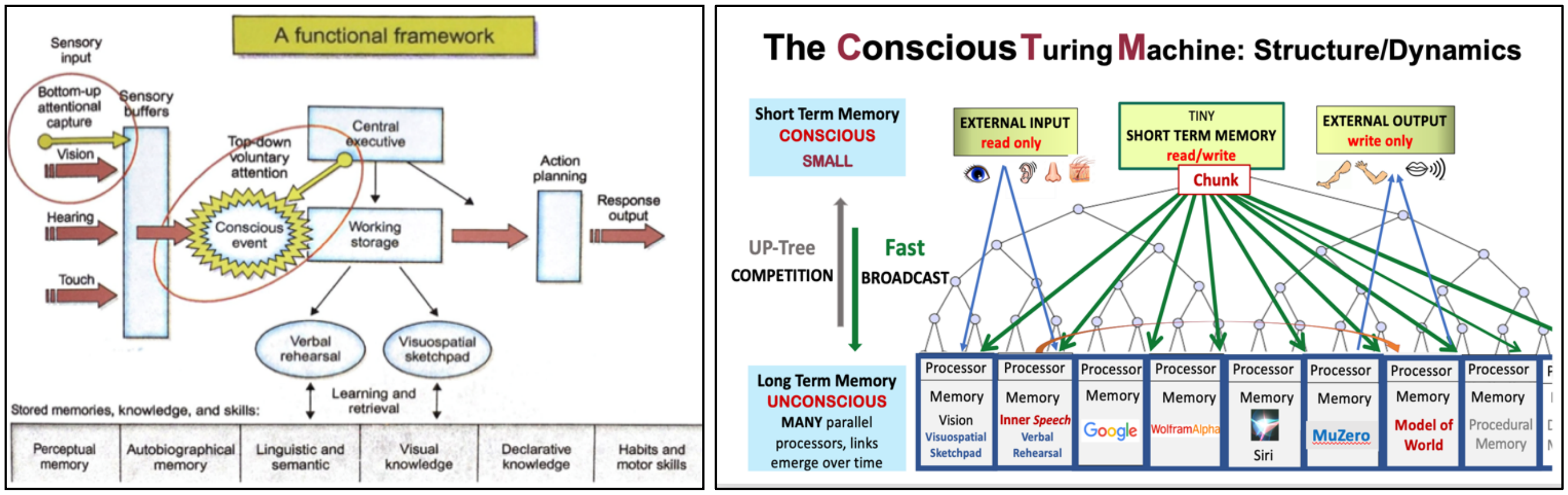

*Figure 4 Baars' **GWT** model (l); **CTM** (r).*

Key features of the **CTM** model and its dynamics resonate with properties of consciousness that (Dennett D. C., 2018) outlines:

> [Neither] a Master Scheduler, nor a Boss Neuron, nor a Homunculus or Res Cogitans [govern the transitions of our conscious minds]. [What governs] must be a dynamical, somewhat competitive process of contents vying for fame, for cerebral celebrity ... or relative clout against the competition. What determines the winners? Something like micro-emotions, the strength of positive and negative valences that accompany and control the destiny of all contents, not just obviously emotionally salient events such as obsessive memories of suffering or embarrassment or lust, but the most esoteric and abstract theoretical reflections.

Although inspired by Baars' **GWT** architecture, the **CTM** integrates features essential for its *feeling of consciousness*. This is the focus of the next chapter.

# 2 The Feeling of Consciousness

While **CTM** is **consciously aware** *by definition* of the **conscious content** broadcast from **STM** (section 1.1.3), this and related notions formalized in the **CTM** are just definitions. Their reasonableness lies in explanations derived from the model that draw confirmation at a high level with the psychology and neuroscience literature.

Here we consider the *feeling* of consciousness in the **CTM** and summarize arguments given in (Blum & Blum, 2021) about how this feeling is generated in the **CTM**.[39] In Chapter 3 we will discuss how the **CTM** provides high level explanations for a range of phenomena generally associated with consciousness. Section 3.5 on **Dream Creation** is particularly relevant to the discussion on the feeling of consciousness.

We argue that the *feeling* of consciousness in **CTM** is a consequence principally of its extraordinarily expressive multi-modal inner language, **Brainish**,[40] coupled with **CTM**'s architecture, certain special processors, and **CTM**'s **predictive dynamics** (cycles of prediction, feedback and learning):

1. **Brainish.** The multimodal Brainish language describes the sensory world as it is perceived. This perception consists of **gists** in the multimodal language of sensations. Its words include gists for odors (the odors as they are perceived by the nostrils), pains (the terribly unpleasant sensations of pains), faces (what one sees when looking at someone's face), and so on. Dreams are important because they show that gists can perfectly express the world when the **CTM** has neither **input** nor **output**.

   To the extent that the **CTM** is successful in the world, that success owes much to the fact that **Brainish gists** give **CTM** its knowledge of the world. Gists are short multi-modal encodings that encapsulate important features of **CTM**'s **inner** and **outer worlds** (definitions in 33 below). A sequence of gists can be a kind of multisensory movie (see **the high level story** in section 1.3). As we shall see, gists are crucial for giving the **CTM** its sense of being alive in the world.

2. **Architecture.** This includes the **Up-Tree** *competition* to gain access to **STM** (section 1.3) and subsequent **Down-Tree** *broadcast of the winner* to all **LTM** processors, particularly all processors that play a special role in generating the *feeling* of consciousness.

3. **Special Processors.** We single out a few (such) processors that have specialized algorithms built into them at birth:

   a. The **Model-of-the-World processor** constructs **models** of **CTM**'s worlds based on information it gets either directly from the environment or from stored possibly modified inner memories.[41] The **Model-of-the-World processor** also has direct output maps to **CTM** actuators, and that some are labeled by past experience, others not.[42] We define **CTM's *inner world*** to be the rough approximate simplified model **"CTM"** that the **Model-of-the-World**

---

[39] Integrated Information Theory (**IIT**) defines a measure of consciousness **PHI** (Tononi, 2004) which roughly speaking measures the amount of feedback in a system. The **CTM** has positive **PHI**, but we wonder if having consciousness according to this measure implies that the entity has the "feeling of consciousness".

[40] As will be discussed in section 3.5, dreams help demonstrate the enormous power of Brainish to generate *feelings* such as the *feeling of consciousness*.

[41] The **Model-of-the-World** processor's **inner memories** are memories stored in the processor itself or memories gotten from other processors via links.

[42] Can you wiggle your ears? What are you doing in your head when you try to do it? To the extent that the human brain is a **CTM**, the human typically decides this by wiggling her ears in her **Model-of-the-World** and looking in a mirror to see (or feeling her ears with her hands to tell) if her ears wiggle. This assumes that the ears in the **Model-of-the-World**, having already been linked to the ears by their sense of touch and hearing, are linked to the actuators of the ear muscles, if any.

**processor** creates of the **CTM**. We define **CTM**'s ***outer world*** to be the model it creates of the **environment**. Importantly the **Model-of-the-World processor** tags parts of its **models of the world** (**inner** and **outer**) with *labels and descriptions* annotated in **Brainish** gists with sensations they can have and actions they can perform.

b. The **Inner Speech processor** extracts whatever speech is encoded in the gist broadcast by **STM** and sends it to the *same locations* that the **Input map** sends gists of **outer speech**, the latter being the speech gists created by the **Input maps**. The **CTM** uses **inner speech** to recollect its past, predict its future, and make plans. The gists of **inner speech** (such as occur in talking to oneself or the talking and hearing in a dream) are nearly indistinguishable from the gists of **outer speech**. In humans, inner speech sounds so much like outer speech that it can be difficult, as in disorders like schizophrenia, to distinguish between Inner and outer speech (Rosen, et al., 2018).

c. **Inner Vision**, **Inner Hearing** and **Inner** (tactile) **Sensation** processors map whatever images, sounds and sensations are broadcast from **STM** to whatever locations the Input maps send outer scenes and outer sensations. The gists of inner vision can be barely distinguishable from the gists of outer vision, the visual gists created by the Input maps. **CTM's** memories and predictive abilities enable **CTM** to create the inner images, sounds and sensations that **CTM** uses to generate imaginings and dreams (section 3.5). To thwart schizophrenic hallucinations, the human brain distinguishes inner images from outer images. The brain has various tricks for doing this, one being to make dreams hard to remember.

These processors inform the "eyes" and "skin" in the **Model-of-the-World** to "see" whatever the **CTM** recalls from visual memory and to "tactilely sense" whatever **CTM** recalls from sensory memory - or creates with its prediction algorithms. These "eyes" and "skin" are **CTM**'s ***mind's eye*** and ***mind's skin***.

We consider these processors, together with the **Inner Speech processor**, as constituting an ***Inner generalized Speech*** **processor**.

Together with the **Model-of-the-World processor**, the **Inner generalized Speech processor** enables **CTM** to recollect its past, predict its future, and make plans.

We emphasize that the **CTM** does not consciously experience the environment directly. The **CTM** sees, hears and senses tactilely what is in its **models of the world**, since inputs go directly to **LTM** processors not (directly) to **STM**.

4. **Predictive Dynamics**. Additionally, we argue that **CTM's** continuous cycling through prediction, feedback and learning (section 0) together with the **stream of consciousness** (section 1.1.3), play a role in **CTM'**s feelings of consciousness (see (James, 1890)). The feelings are enhanced by (parallel) predictive dynamics in **CTM**'s **Model-of-the-World** where planning and testing is constantly carried out, often before action is taken by the **CTM**. Positive feedback gives **CTM** an indication that it understands what is going on; negative feedback – unless it is about something that could not have been predicted such as an unexpectedly loud noise – gives **CTM** evidence of something that it did not know or understand.

We also add, but do not develop here:

5. A minimal (general) ability to think and make plans, and
6. **Motivation** (**= energy + drive**) to make a plan and then pursue it.

We now return to the **Model-of-the-World processor** to describe one of its central tasks, that of tagging various constituent parts of its models as either **self** or **not-self** (else unknown).

How does the **Model-of-the-World processor** determine what is or is not **self**? If the broadcast of a chunk (a **CTM** thought) is immediately followed by an actuator carrying out an action in the environment– and that same thought leads to the same action consistently and repeatedly – then that indicates the actuator is part of self.

For example, suppose the **CTM** wishes to lift a rock in the environment "with the power of thought". This means that a processor attempts to lift the rock in the **real world** (i.e., environment) - without using any of its known actuators – merely by lifting the rock in the **Model-of-the-World** directly. If that request succeeds in the real world, meaning the **CTM** detects that the real world rock got lifted this way, then the **Model-of-the-World processor** labels the rock and the thought (gist) that lifts it as **self**.

The **Model-of-the-World processor** has additional important jobs that give the **CTM** its ***sense of self***, including creating imaginings; creating maps of its environment; registering movements in its environment; helping to plan actions in the environment; helping to predict the actions of **self** and **not-self** in the environment; and correcting predicted actions of **self**.

We emphasize that the **"self"** in the **models of the world** is (appears to be) the **self** in the real world as far as the **CTM** is concerned. Brainish gists are much more than English labels: they are descriptions of conscious thoughts every bit as convincing as the descriptions that appear in dreams (see section 3.5).

When (through broadcasts) the **CTM** detects itself thinking about its own consciousness, the **Model-of-the-World** processor tags the "**CTM**" in its models as "**conscious**".

We now look at why the **CTM** considers itself conscious. It cannot be because the **Model-of-the-World processor** or any other processor feels it is conscious, as processors are just machines running algorithms – and (such) machines have no feelings. We propose that **CTM** as a whole ***feels* conscious**, as the term is normally understood, as a consequence in part of the fact that the **Model-of-the-World processor** views the "**CTM**" in its model of the world, as "conscious", and that this view is broadcast to all processors.[43] Here, **"CTM"** is a simple learned representation of the much more complex **CTM**.

Our argument for the feeling of consciousness aligns with Michael Graziano's argument for m-consciousness in the **Attention Schema Theory** (**AST**). See (Graziano, Guterstam, Bio, & Wilterson, 2020).

Carpe Diem Comics

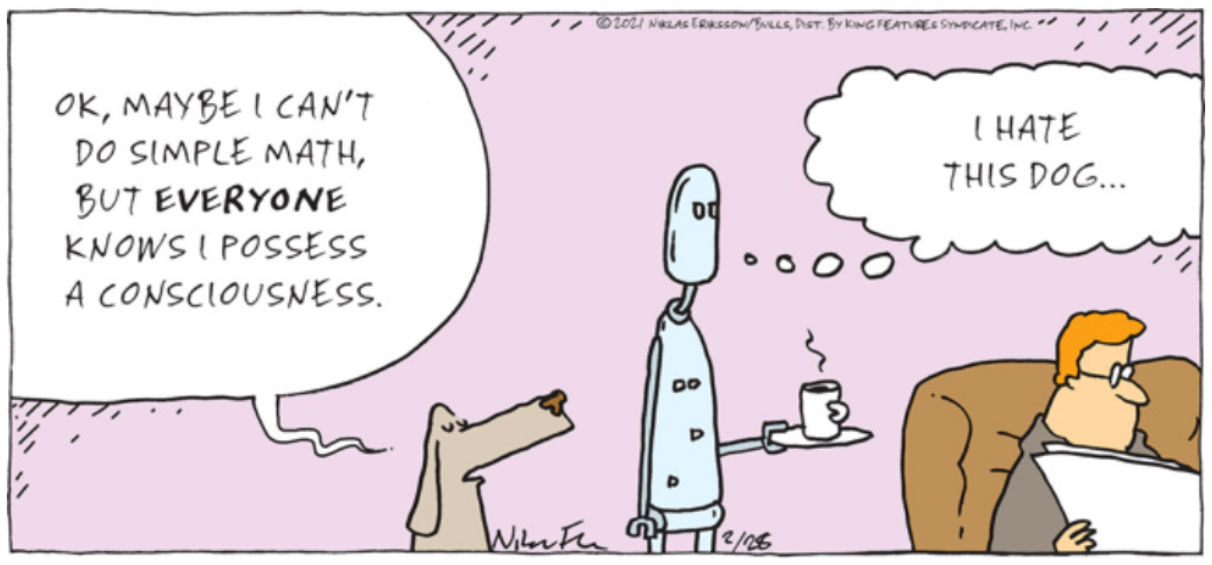

[43] Shimon Edelman says (personal communication) "This [explanation for the feeling of consciousness in the **CTM**] reminds me strongly of the explanation offered by (Metzinger, 2004)".

# 3 High Level Explanations

We now explore how **CTM** might experience a variety of phenomena generally associated with consciousness. We believe that our explanations, *derived from the model,* provide a *high level* understanding of how conscious experiences might be generated. These draw confirmation from consistencies with the psychology and neuroscience literature, again at a *high level*. Indeed, we argue that a very simple well-defined device, the **CTM**, supports *high level* explanations for a great many phenomena whose explanations are otherwise extremely complex.

Previously (Blum & Blum, 2021) we explored explanations for the feelings of pain and pleasure in the **CTM**. Here we consider additional phenomena, again from the perspective of the **CTM**. We start with three examples related to vision (blindsight, inattentional blindness, and change blindness), then follow with a discussion of dreams, free will and altered states.

## 3.1 Blindsight

*Blindsight* provides a striking example of the difference between conscious and unconscious awareness (Striemer, Chapman, & Goodale, 2009). In blindsight, the person does not *consciously* see the outer world. When asked to fetch something across a cluttered room, a typical response is “But I cannot see.” Nevertheless, the person responds adeptly if cautiously to the request. What is going on?

In the **CTM,** visual **Inpu**t goes directly from the **vision sensors** to a subset of **LTM** processors that process visual input. But in the blindsighted **CTM**, due to some malfunction, perhaps a break in the **Up-Tree** or some other inability for the **Vision processors** to enter chunks competitively into the competition, this information does not get up to **STM** and hence does not get globally **broadcast**. For this reason, **CTM** is *not* **consciously aware** that it can see. However, information can still be communicated between (unconscious) processors via **links**. So visual information received by the **Vision processors** can be sent through links to the **Walk Processor** that controls the leg actuators.

At a high level, this explanation is consistent with explanations of blindsight in humans given by (Ajina & Bridge, 2018).[44]

## 3.2 Inattentional Blindness

*Inattentional blindness* occurs when an individual fails to perceive a visual stimulus that is in plain sight. It is “the failure to notice the existence of something unexpected when attention is focused on some other task” (Jensen, Yao, Street, & Simons, 2011).

For example, in the famous *selective attention test* of (Chabris & Simons, 1999), viewers of the “invisible gorilla” film were asked to “count how many times the players wearing white shirts pass the basketball” (*Figure 5*). Nearly all viewers gave close to the correct number (15), but were stunned when asked, “Did you see the gorilla?”

---

[44] (Ajina & Bridge, 2018) assert that when the primary visual cortex (V1) is damaged impairing conscious vision, blindsighted individuals still have functional connections between the lateral geniculate nucleus (which receives input from the retina and projects this information to V1) and hMT+ (the cortical area that detects motion). This functional connection was absent in V1 impaired patients without blindsight.

In blindsighted individuals, some retinal input travels (unconsciously) directly to hMT+ bypassing V1. We view it likely that this information is transferred (unconsciously) to, and acted on, by the motor cortex.

**Selective attention test:**

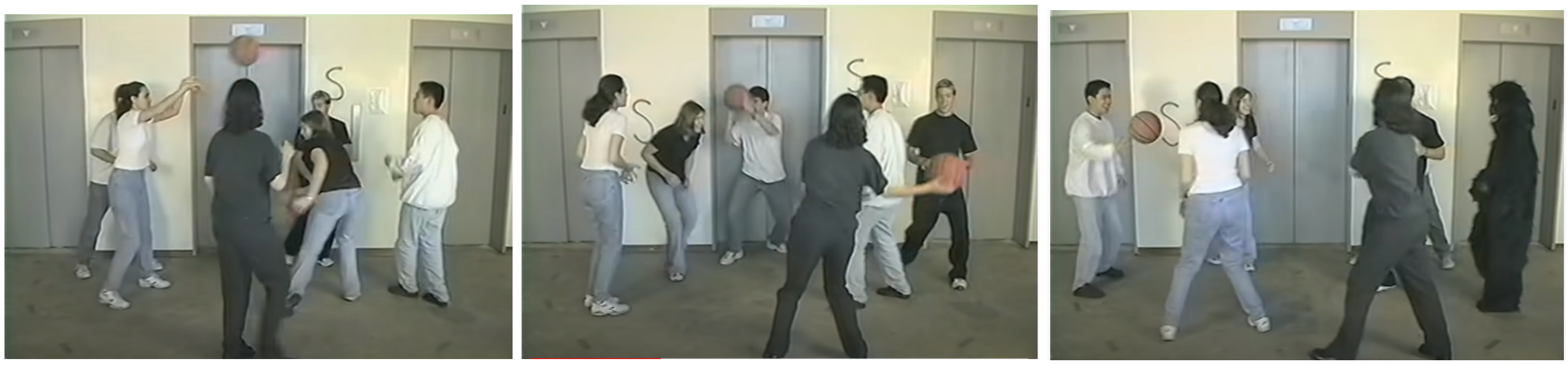

*Figure 5 Screen shots from video, "The original selective attention task" (Chabris & Simons, 1999).*

What is going on?

Let's suppose the **CTM** is viewing the film. The **Input** query about the white shirted players gains access to **STM** and is then immediately broadcast to all **LTM** processors. To carry out the task, **CTM**'s **Vision processors** assign high intensities to white shirted gists and very low (possibly zero) intensities to anything black. The chunk with the "gorilla" gist has little chance to enter **STM**. The **CTM** does not consciously see the gorilla.

The **CTM** explanation of inattentional blindness reduces to the differential intensities given to gists, lower intensities given to irrelevant ones, and the competing advantages of chunks with higher intensities.

According to simulations performed by (Dehaene & Changeux, 2005), during certain "ignited" states, "spontaneous activity can block external sensory processing." They relate this blocking to the cause of inattentional blindness. In our view, blocking the "sensory processing" in human brains of black objects is roughly equivalent to the **CTM** dramatically lowering the intensity of black gists in chunks, thus lowering the chances of those chunks to enter **STM**. The effect of differential intensities in the **CTM** is also consistent with theoretical implications that inattentional blindness in humans "can serve as a filter for irrelevant information." It may also filter out unexpected events (Jensen, Yao, Street, & Simons, 2011).

## 3.3 Change Blindness

*Change blindness* occurs when individuals fail to notice large changes in pictures or scenes (Rensink, O'Regan, & Clark, 1997). It is "the failure to notice when something has changed from one moment to another" (Jensen, Yao, Street, & Simons, 2011).

An instructive example is the (Test Your Awareness : Whodunnit?) video. A detective enters a murder scene proclaiming, "Clearly somebody in this room murdered Lord Smythe" and immediately interrogates each suspect in turn. The maid proclaims, "I was polishing the brass in the master bedroom," the butler "I was buttering his Lordship's scones," and Lady Smythe "I was planting my petunias in the potting shed." Enough information for the clever detective to solve the murder on the spot.[45]

But why didn't we notice the many incongruous scene morphs between the beginning screen shot and the end? (See *Figure 6*.)

[45] Detective: "Constable, arrest Lady Smythe." Lady Smythe: But, but, how did you know?" Detective: "Madam as any horticulturist will tell you, one does not plant petunias until May."

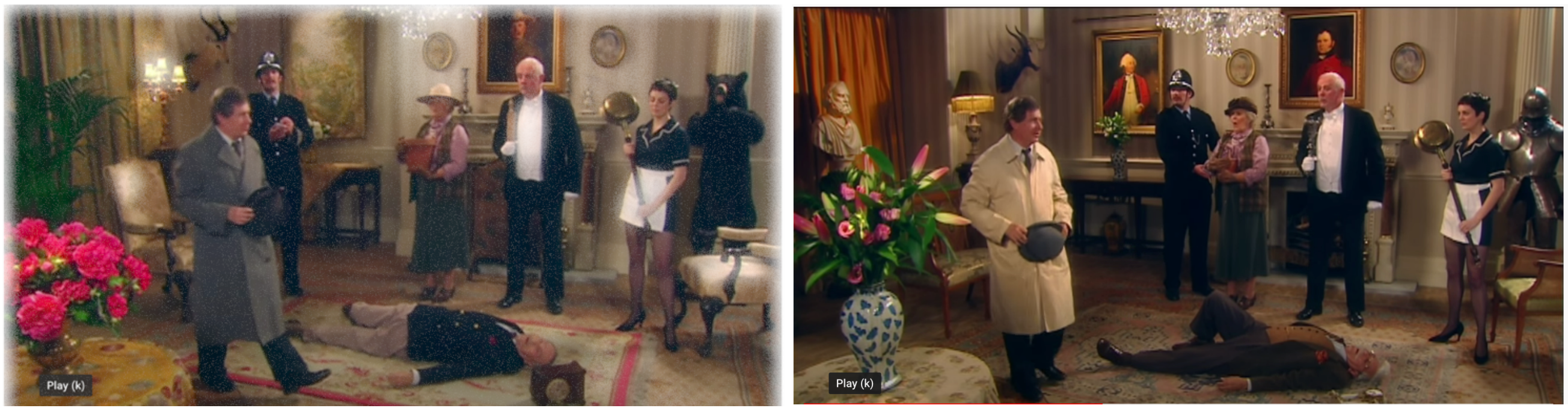

*Figure 6 Beginning and ending screen shots from video (Test Your Awareness : Whodunnit?, 2008).*

From the perspective of the **CTM**:

In viewing the *Whodunnit* video (*Figure* 6), **CTM** has the impression of seeing the whole but doesn't notice the changes that take place as trench coat, flowers, painting, and so on are replaced by variants. That is because:

1. The filming is cleverly staged so that there are cuts from the whole scene to the suspects (e.g., the maid alone), eliminating transitions that show the dark trench coat replaced by the white one, the bear replaced by the suit of armor, the rolling pin by the candelabra, the dead man now with a change of clothes and raised leg, and so on. The video **Input** never signals **CTM**'s **Vision processor** that the "scene" has been modified.
2. And more importantly, the same gist describes both the beginning and ending scenes equally well: "The living room of a mansion with detective, butler, maid, others, and a man apparently dead on the floor."

Under these conditions, the **CTM** experiences *change blindness*.

Again, the **CTM** explanation is consistent with literature on change blindness in humans. For example, according to (Jensen, Yao, Street, & Simons, 2011) confirming earlier work of (Rensink, O'Regan, & Clark, 1997):

> "Given that change detection requires adequate representation of the pre- and post-change scenes as well as a comparison, any task characteristics that influence the richness of the representation or the tendency to compare representations should affect detection. The semantic importance of the changing object appears to have the biggest influence on the likelihood the subject will attend, and therefore notice, the change."

In other words, detecting change in the *Whodunnit?* video would have required significant changes in the gists describing the beginning and ending scenes. But size limitations on conscious content (and the clever scene transitions) caused the high level descriptions to be essentially the same.

## 3.4 Illusions

Inattentional Blindness and Change Blindness might be considered examples of illusions.

The **CTM** is consciously aware (by definition) of the gists (in chunks) that are broadcast from **STM**. (Those gists reached **STM** from **LTM**. **LTM** got them directly from **sensors** via **Input maps**, from other **LTM** processors through **links**, and from **STM** by **broadcasts**.). The gists are stored in **LTM** memories for many reasons, one being for use by **SEAs**, another to supply the processors' high level stories (section 1.3) such as those that occur in dreams.

In **CTM**, the **stream of consciousness** is the sequence of gists broadcast from **STM** (section 1.1.3). Each visual gist at each moment gives the **CTM** the sense that it sees the entire scene before its eyes, though in truth it sees at most a tiny fraction of the scene. The illusion of the whole has several explanations, the main one being that a multi-modal Brainish gist can describe a hugely complex scene like "I'm standing before a Japanese style garden containing a brook, path, bridge and trees." Could that gist contain the details of a 12 million pixel photograph from an iphone camera, which is what it *feels* like we are seeing? The illusion of the whole is a consequence of the highly suggestive information in a gist. The **CTM** conjures up the scene in a kind of magic act. Keith Frankish (Frankish, 2016) calls this the *illusionism theory of consciousness*.

## 3.5 Dream Creation

Dreams are the ultimate illusions. Some people claim not to dream, but most do (Herlin, Leu-Semenescu, Chaumereuil, & Arnulf, 2015). Their dreams may be visual, auditory, tactile, etcetera. They are often related to emotional processes (Freud S. , 1900), (Scarpelli, Bartolacci, D'Atri, Gorgoni, & De Gennaro, 2019). They can express great pain and fear (nightmares), or great pleasure (as in flying dreams). One can feel crippling pain in the leg and wake up to find that the pain is completely illusory: there is no pain at all. One can be lying face down and wake face up.

In the **CTM**, a built-in **Sleep processor** keeps track of time, habits, day/night etc. and has internal algorithms to monitor the need for sleep. If and when the **Sleep processor** determines that sleep is needed, it takes control by raising the intensity of its own chunks enough to get them into **STM** and to keep other chunks out. This has roughly the same effect as lowering the intensities of chunks from other **LTM** processors. It also *blocks* or *greatly reduces* the intensity of various inputs (eyes and ears), and it *blocks signals that activate outputs* (such as to limbs). The **CTM** sleeps. This is the **sleep state**.

The **Sleep processor** continuously monitors the need for sleep, and as that need diminishes, reduces the intensity of its own chunks proportionately. This eventually permits **dream gists** (in chunks) to reach **STM**. This is the **dream state**. Finally, when the **Sleep processor** releases its choke hold on inputs and outputs, the **CTM** wakes up. That's in the **CTM**. In humans, non-REM and REM sleep can alternate several times before awakening (Vyazovskiy & Delogu, 2014).

When **CTM** is in the dream state, a processor acting as **Dream Creator** becomes active (that is, starts getting its chunks into **STM**). The gists in these chunks contain kernels of ideas (typically based on earlier **CTM** activities, concerns, imaginations). When these chunks are broadcast, *all* processors, including those that play key roles in the feeling of consciousness, receive those broadcasts and compete to respond. This gives the **CTM** the same sense of being alive while in the dream state as when it is awake.

The **Dream Creator** and the other processors take turns interacting back and forth. The conversation – the back and forth interaction – between **Dream Creator** and the gamut of processors is the sequence of gists that constitutes the **Dream**. This sequence is the **dream stream** of consciousness.

The **Dream** essentially stitches together this sequence of chunks to produce a **dream stream of consciousness**, aka **inner movie**, that

1. sees, hears, and senses the dream world, and
2. affects what appears in that dream world.

Such an (interactive) inner movie displays a range of sensory inputs (images, smells and sounds) and generates a range of actions.

When the **CTM** is asleep but not dreaming, most processors cannot get their chunks into **STM.** Exceptions include detectors of especially loud noises, and the **Sleep processor** itself. The **Sleep processor's** chunk in **STM** blocks most other processors' chunks from reaching **STM**. By design, it holds an empty gist, so the **CTM** is not conscious or barely so.

After the **CTM** leaves the **sleep state** to enter the **dream state,** a fraction of **LTM** processors such as the **Inner Vision processor** can get their chunks into **STM**. Thus, while dreaming, the **CTM** is conscious and can experience events vividly.

Dreams demonstrate the power of **Brainish gists**. What **CTM** sees, hears, feels and does in a dream are necessarily *fabrications* by processors that can recall, modify, and submit creations to the competition for **STM**. These fabrications are realistic because they use the same gists that are generated while awake. Thus, dreams generate the sense of a realistic world even while **CTM** is completely divorced from external inputs. As a consequence, they can appear so realistic that for **CTM**, as for humans (Corlett, Canavan, Nahum, Appah, & Morgan, 2014), it may become hard to distinguish dreams from reality. (This problem is avoided in humans if dreams are hard to remember.)

Confirmation: In *When Brains Dream*, (Zadra & Stickgold, 2021) refer to research by (Horikawa, Tamaki, Miyawaki, & Kamitani, 2013) demonstrating that in humans, the same neural pattern of activity occurs when one sees a face, brings the face back from memory, or when the face appears in a dream. They also point out that in REM sleep, the activation of the motor cortex in a dream, when one has the sensation of movement, is the same activation as when awake.

As discussed earlier (Chapter 2), key processors such as those for **Inner Speech**, **Inner Vision, Inner Sensations,** and **Model-of-the-World** play special roles in generating the "feeling of consciousness" in **CTM**. These processors play similar roles when **CTM** is dreaming.

Here are some examples of how processors help with dream creation:

- The **Inner Speech processor** culls the inner speech from the multi-modal gists broadcast from **STM** and sends that speech to the same processor that receives outer speech.[46] This process causes speech in dreams to sound like outer speech. The **Inner Vision** and **Inner Sensation** processors help in a similar way with dream creation.
- The **Model-of-the-World processor** predicts the effect that **CTM**'s actions will have in its (inner and outer) world. It does this from the effect of those actions in its models of the world. The **Dream Creator** can use this same prediction machinery to create dreams.

Dreams also enable the **CTM** to test itself in unknown and possibly dangerous situations. In both humans and **CTM**, dreams can be laboratories for experimenting with various possible solutions.

However, unlike what occurs in waking consciousness, inconsistencies are *more likely* to occur unnoticed in dreams than while awake since the **CTM's** "consistency checkers" in its **Model-of-the-World** are not getting input from the environment. Hence the **CTM** can fly in its dreams

(Zadra & Stickgold, 2021) assert that in humans, "Dreams don't replay memories exactly; they create a narrative that has the same gist as some recent memory and could have the same title." They note that "REM sleep provides a brain state in which weak and unexpected associations are more strongly activated than normal strong associations, explaining how it aids in finding the remote associates and perhaps explaining the bizarreness in our REM sleep dreams."

[46] Spoken speech is not heard as speech until the appropriate links between speaker and hearer are created. These links would appear most likely early in the life of the **CTM**.

## 3.6 Altered States of Consciousness

Under psychedelics or meditation, humans can experience altered states of consciousness ranging from a heightened sense of awareness to dissolution of self (feelings of being "one with the world"). We agree with (Bayne & Carter, 2018) that these are *states*, not levels, of consciousness. We disagree, however, with their assessment that the global/global neuronal workspace theories are too simple to explain these altered states. Indeed, the beauty of those theories lies in the significant understanding that comes of their simplicity.

To show how the **CTM** might experience a simple form of *dissolution of self*, we start by describing a **Mindful Meditation processor (MMp)** that a **CTM** might have.

The conscious decision to meditate would be the concern of an **MMp** that creates and submits a sequence of chunks to the competition for **STM**. Through repeated practice, this processor gains strength and increases the intensity of its chunks. It can be surprisingly difficult for the **MMp** to keep other chunks from entering **STM**: the difficulty is not in the sense of lifting a heavy weight or proving a difficult theorem, but in the sense of demanding focused concentrated attention and practice. (Rathi, 2021) explains how a human, using the *Mantra* meditation technique, accomplishes this.

When the **MMp** is successful, its chunks get into **STM** and are broadcast. Those broadcasts generally contain feedback that other processors use, through their **Sleeping Experts Algorithm**, to *hush* their own self-evaluations (section 0). Thus during successful meditation, the **MMp**'s chunks get the lion's share of time in **STM**.

Additionally, during successful meditation, chunks that get communicated via **links** from all processors except the **MMp** get hushed – by the incoming broadcasts from **MMp** – and thus processors are unlikely to pay their usual attention to the chunks they receive through links.[47]

This "hushing" or diminishing of functional connectivity is observed in studies on effects of psychedelics and meditation. For example, brain imaging and electromagnetic studies on effects of certain psychedelics (psilocybin) suggest that the dissolution of self ("ego-dissolution") is due to "disintegration" of functional connectivity (Calvey & Howells, 2018).[48] This decreased connectivity accounts in part for the sense of dissolution of spatial boundaries, which in turn leads to the feeling of being "one with the world".

---

[47] During successful meditation, the hushing of chunks to be communicated via links diminishes link communication. This in effect diminishes the **Model-of-the-World** processor's ability to communicate to others what is **self** and what is **not-self**.

Neuroimaging studies on various forms of meditation from distinct traditions share some common neural correlates, see (Millière, Carhart-Harris, Roseman, Trautwein, & Berkovich-Ohana, 2018). Importantly, the latter report that in several forms of meditation there is "attenuation for either activity or functional connectivity" in the medial prefrontal cortex and in the posterior cingular cortex, key nodes of the so-called default mode network (DMN). The DMN is active when a person is daydreaming or mind-wandering. It is also active when a person is thinking about others or themselves, remembering the past, or planning for the future, see (Buckner, Andrews-Hanna, & Schacter, 2008) and (Lieberman, 2013). Thus attenuation of functional connectivity in these areas may also account for dissolution of self.

[48] Referencing (Carhart-Harris, et al., 2016), (Calvey & Howells, 2018) suggest that this "disintegration" is due at least in part to decreased connectivity between the parahippocampal place area (PPA) and the retrosplenial complex (RSC). According to (Epstein, 2008), the PPA is "concerned with representation of the local visual scene" while the RSC is "more concerned with situating the scene within the broader spatial environment."

## 3.7 Free Will[49]

It matters not how strait the gate,
How charged with punishments the scroll,
I am the master of my fate,
I am the captain of my soul.
- William Ernest Henley (1875)

**The Problem of Free Will** is ancient. It appears in Lucretius [*De Rerum Natura,* 1st century BC]:
"If all movement is always interconnected, the new arising from the old in a determinate order – if the atoms never swerve so as to originate some new movement that will snap the bonds of fate, the everlasting sequence of cause and effect - what is the source of the free will possessed by living things throughout the earth?" (Lucretius & Ferguson Smith (translator), 1969).

**The Paradox of Free Will** is captured by Dr. Samuel Johnson's (1709-1784) observation (Boswell, 1791):
"All theory *is against* the freedom of the will; all experience *is for it*."

Stanislas Dehaene (Dehaene S. , Consciousness and the Brain: Deciphering How the Brain Codes Our Thoughts, 2014) supplies a contemporary voice:
"Our brain states are clearly not uncaused and do not escape the laws of physics – nothing does. But our decisions are genuinely free whenever they are based on a conscious deliberation that proceeds autonomously, without any impediment, carefully weighing the pros and cons before committing to a course of action. When this occurs, we are correct in speaking of a voluntary decision – even if it is, of course, ultimately caused by our genes [and circumstances]."

We add to Dehaene: computation takes time. To make a decision, a **CTM** evaluates its alternatives, and such an evaluation takes time. During that time the **CTM** is free, can use its **inner speech** to tell itself it is free, can *feel* free, to choose whichever outcome it deems (i.e., computes) best.

The **TCS** perspective directly addresses **The Paradox of Free Will** and informs our definitions of **Free Will** and the ***feeling of* free will:**

> First, **Free will** is the ability to make a decision that violates the laws of physics. As that is impossible,[50] no one has Free Will.
>
> Second, the ***feeling* of free will** is the ***conscious** knowledge*, when one has it, that when faced with a choice, one can compute the consequences of different courses of action – or as much of those consequences as is possible with the available resources (time, space, computational power, and information) – and choose whichever course of action best suits one's utility and goals. That *feeling* of free will *is* something one can have.

This definition incorporates both *predictive dynamics* (compute the consequences of different courses of action) and *resource constraints* (time, space, computational power, and information).

For example, consider a **CTM** that is called on to play a given position in a game of chess. Different processors suggest different moves. The **CTM**'s main chess-playing processor (assuming one exists, else a processor that has a "high level" view of the game) indicates, by broadcast of a chunk from **STM**, that it recognizes it has a choice of possible moves and that the decision which move to make merits a careful look at the consequences of each move. At this point, faced with a selection of possible moves but not yet having evaluated the consequences of those moves, the **CTM** is free to choose whichever move it reckons best – *within* the time constraints.

---

[49] See also, (Blum & Blum, A Theoretical Computer Science Perspective on Free Will, 2022).

[50] Violating the laws of physics is clearly impossible in a deterministic world. It is also impossible in a probabilistic world in which all actions are random selections from a well-defined probability distribution.

Will the **CTM** *feel* that it has **free will**?

1. Consider the moment that the **CTM** asks itself "What move should I make?" meaning this question has risen to the **STM** stage and, through broadcast, has reached the audience of **LTM** processors. In response, a number of those processors submit suggestions to the competition. The winner of the competition reaches the stage and gets broadcast. Because gists are short, any such broadcast is short and therefore reasonably articulable.
2. The continued back and forth comments, commands, questions, suggestions and answers that appear in **STM** are globally broadcast to **LTM**. They give the **CTM** ***conscious*** *knowledge of its control*: If the **CTM** were asked how it generated a specific suggestion, i.e., what thinking went into making that suggestion, its processors, with the help of its **Inner Speech** processor, would be able to articulate the fraction of conversation that reached the stage (though perhaps not much more than that in the short term).
3. Many **LTM** processors compete to produce the **CTM**'s final decision, but **CTM** is only *consciously aware* of what got into **STM**, which is not all of what was submitted to the competition. Moreover, much of **CTM**, meaning most of its processors, are not privy to the unconscious chatter (through links) among processors. To the **CTM**, enough is consciously *unknown* about the process that the decision can appear at times to be *plucked from thin air*. Even so, although **CTM** does not consciously know *how* its decisions were arrived at, except for what is in the high level broad strokes broadcast by **STM**, it *knows* that its decisions came from inside itself. The **CTM** can rightly take credit for making its decisions (after all, they did come from inside the **CTM**), can explain some of them with high level stories (see section 1.3), and as for what it cannot explain, it can say "I don't know" or "I don't remember." It is the knowledge that there are choices, that it (the **CTM**) has knowledge of those choices – and that it has ignorance as well – that generates the *feeling* of free will. Deterministic or not, the *experiential feeling* is one of free will.

How important is randomness for this explanation of the *feeling* of free will? Notice that *no quantum physics* is required in the **CTM** for the above explanation. The only randomness is that of the **coin-flip neurons** in the **Up-Tree competition** and whatever randomness, if any, the processors use in their probabilistic algorithms. It can be shown, moreover, that the above argument for the *feeling* of free will still applies for a *completely deterministic* **CTM**, e.g. one that uses pseudo-randomness, i.e., the output of a pseudorandom generator that has been provided with a (short) random seed in place of true randomness. It follows – and we expect this will be a source of contention – that even in a *completely deterministic* world, the **CTM** will *feel* it has free will.

What is the significance of the *time delay* (section 1.4) between chunks being submitted by the unconscious processors into competition and **CTM** becoming consciously aware of the winner? Some such delay is necessary to select one of **N** chunks for broadcast (plus the single tick for the broadcast). This delay is consistent with the (Libet, 1985) experiments showing delay between unconscious decisions and conscious actions.

About the Libet experiments: there has been substantial controversy over the interpretation of the Libet experiments for the existence or not of free will (Gholipour, 2019). For example, some research shows that the measured delay may be due to effects of stochastic fluctuations (Schurger, Sitt, & Dehaene, 2012), some argue that the distinction between deliberate or arbitrary decisions have to be taken into account (Maoz, Yaffe, Koch, & Mudrik, 2019). Other research on volition qualitatively corroborates the earlier results, e.g., (Fried, Mukamel, & Kreiman, 2011) and (Haggard, 2011). While those results present insight into brain dynamics, we do not view them as providing arguments for or against free will.

# 4 Summary

We consider consciousness from the perspective of theoretical computer science (**TCS**), a nonexperimental area of mathematics. Inspired by Alan Turing's simple yet powerful model of a computer, the Turing Machine (**TM**), and by Bernard Baars' *Theater of Consciousness,* we define a computational model of consciousness, the **Conscious Turing Machine** (**CTM**).

The **CTM** is *defined formally* (Chapter 0) as a **7**-tuple, **< STM, LTM, Up-Tree, Down-Tree, Links, Input, Output >**. The theory includes a precise definition of George Miller's informally defined **chunk**, and a precise definition of a **competition** for deciding which of the ($10^7$ or more) **Long Term Memory (LTM)** processors gets access to **Short Term Memory (STM)**.

Bi-directional **links** between processors that *emerge* in the life of the **CTM** enable conscious processing to become unconscious. Links are also especially crucial for the "global ignition", described by (Dehaene & Changeux, 2005) in their **Global Neuronal Workspace Theory (GNWT)**, that re-enforces and sustains conscious awareness. **Input/Output** maps enable communication between the **CTM** and its **environment**. Other features of the model can be found in (Blum & Blum, 2021).

In particular, we argue (Chapter 2) that the *feeling* of consciousness arises in **CTM** as a consequence of:

1. the global workspace *architecture,* which enables all processors, including *especially those that are particularly responsible for consciousness* – **inner Speech**, **Inner Vision**, **Inner Sensations** and **Model-of-the-World** – to be privy to the same (conscious) content of **STM**,
2. the *expressive power* of **CTM**'s multi-modal inner language *Brainish*,
3. the close correspondence between *gists* of outer speech (what we say and hear in the world), outer vision (what we see in the world), and so on, to gists of **inner speech** (what we say to ourselves), inner vision (what we see in dreams), and the like, and
4. *predictive dynamics* = cycles of prediction, feedback, and learning that help **CTM** develop its understanding – its ability to deal with – its environment and inner world.

We argue (Chapter 3) that the *feeling* of free will in the **CTM**, like the *experiences* of illusions and dreams, are direct consequences of **CTM**'s *architecture,* certain *special processors* such as the **Model-of-the-World** *processor* and the ***Inner generalized Speech*** *processor*, the *expressive power of* ***Brainish***, and its *predictive dynamics*.

The paper (Blum & Blum, 2021) and an expanded monograph (Blum, Blum, & Blum, Towards a Conscious AI: A theorectical computer science model of consciousness inspired by cognitive neuroscience, monograph in preparation) cover the topics presented here in considerably more detail, including especially **Brainish**, the **Sleeping Experts Algorithms**, the **Hard Problem** (Chalmers, 1995) for **pain** and **pleasure** and properties of the **LTM** processors.

# 5 Relation to Other Theories of Consciousness

The **CTM** is an abstract computational model designed to consider consciousness from a **TCS** perspective. It is not intended to model the brain nor the neural correlates of consciousness. Nevertheless, the **CTM** is both inspired by, and has certain features in common with, neural, cognitive, and philosophical theories of consciousness.

The **CTM** is directly influenced by Bernard Baars' **GWT** (Baars B. J., 1997), which is supported by (Dehaene & Changeux, 2011), (Dehaene S. , Consciousness and the Brain: Deciphering How the Brain Codes Our Thoughts, 2014) and (Mashour, Roelfsema, Changeux, & Dehaene, 2020) in their investigation of neural correlates of consciousness known as the Global Neuronal Workspace Theory (**GNWT**). We are inspired by David Mumford's 1991 work on the computational architecture of the neocortex (Mumford, 1991), which we view as an early proposal for **GNWT**.

Like the **LIDA** model of cognition (Baars & Franklin, 2007) and (Baars & Franklin, 2009), **CTM** is architectural. Unlike **LIDA**, which is a more elaborate model of **GWT**, the **CTM** is intended to be a *minimal* model of **GWT** sufficient to explain a wide range of conscious phenomena and, in particular, the *feeling* of consciousness.

*Predictive dynamics* (the ensemble of prediction, feedback, and learning) is an additional key feature of the **CTM**. It is related to the notion of predictive processing (**PP**), see (Lee & Mumford, 2003) (Friston, 2003) (Friston, 2005) (Cleeremans, 2014) (Clark, 2015) (Seth, 2015) (Hohwy & Seth, 2020).

We see a kinship between the **CTM** and the self-aware robots developed by (Chella, Pipitone, Morin, & Racy, 2020). We also see a kinship between the **CTM** and the Global Latent Workspace (**GLW**) proposed by (VanRullen & Kanai, 2021) for deep learning. We view the **CTM** as providing a framework for machine consciousness. The properties of C0, C1, C2 that (Dehaene, Lau, & Kouider, 2017) suggest for machine consciousness map nicely to like properties of the **CTM**.

Our explanation for **CTM**'s "feeling of consciousness" aligns closely with Michael Graziano's Attention Schema Theory (**AST**) (Graziano, Guterstam, Bio, & Wilterson, 2020). As in **AST**, **CTM** is consciously aware of both external and internal events. Basic **AST** is similar to **GWT**: its i-consciousness (i for information) aligns somewhat with **CTM**'s conscious awareness. Full **AST**'s m-consciousness is similar to **CTM**'s "feeling of consciousness".[51]

However, we do not agree with Graziano et al. that **GWT** "leaves unexplained how people end up believing they have subjective experience" i.e., that it leaves an explanatory gap. Instead, we argue that in our model, the feeling of subjective experience arises when "winning chunks" from imaginings and dreams, for example, are received by the same (unconscious) processors that receive chunks directly from the environment via **Input maps**. Additionally, the **Model-of-the-World processor** incorporates the information gotten from the winning chunks (i.e., the **conscious content** of the **CTM**) into its **models of the world**, as appropriate, tagging the "**CTM**" in all its **models of the world** as "conscious". This is similar to Graziano's argument for consciousness in the **AST**. Fuller discussion for the *feeling* of consciousness in the **CTM** is in Chapters 2 and 3, and in (Blum & Blum, 2021).

Philosophically, we align with much of Daniel Dennett's functionalist perspective (Dennett D. C., 1991) except we don't agree with his view that we are the only species to have consciousness (Dennett D. C., 1978) (Dennett D. C., 2019). As for animal consciousness, we agree with (Mumford, 2019) that consciousness is a matter of degree. Here he cites (Merker, 2007) that consciousness does not need a cerebral cortex: it arises from midbrain structures. We would also cite other studies, e.g., (Slobodchikoff, 2012).

[51] Full **AST** has three neural networks (**A** for receiving information, **B** for constructing an attention schema, and **C** for reporting to the outside world) to obtain a system which purportedly thinks it has subjective experience (m-consciousness, m for mysterious).

We do not see the *explanatory gap* (Levine, 1983) between functional and phenomenological consciousness as insurmountable. This viewpoint aligns closely with Baars (see (Kaufman, 2020) interview) and (Dennett D. C., 2016). Indeed, we see the **CTM**'s ability to tag and test features in its **models of the world** as playing a role in the feeling of "what it is like" (Nagel, 1974).

Both **AST** and **CTM** appear to embody illusionist notions of consciousness proposed by (Dennett D. C., 2019) and Keith Frankish (Frankish, 2016). Saying that the feeling of consciousness is an illusion does not deny the existence of that feeling. As a familiar example, the fact that a movie is made up of (many) discrete still images does not affect the feeling of continuity one gets from viewing it. The feeling of continuity is an illusion.

By utilizing existing technology (or apps) to supplement its supply of **LTM** processors (section 0), **CTM** incorporates elements similar to those advocated by (Clark & Chalmers, 1998)'s "extended minds".

Integrated Information Theory (**IIT**), the theory of consciousness developed by Giulio Tononi, (Tononi, 2004) and supported by Koch (Tononi & Koch, 2015), proposes a measure of consciousness called **PHI**, defined using Shannon's information theory that essentially measures the amount of feedback in a system. It is a mechanism's intrinsic ability to influence itself, rather than its input-output information processing, that determines its consciousness.

This is consistent with **CTM**'s *intrinsic* predictive dynamics (of prediction, feedback and learning). Tononi proposes five "axioms" (properties) necessary for any causal system to have consciousness.[52] Given a detailed specification of a **CTM**, one could in principle compute its **PHI** and compare it to the **PHI** of any other precisely defined causal system. It turns out that many causal physical systems have non-zero measures of **PHI**. **IIT** would validate animal consciousness.

With regard to the "adversarial collaboration" between advocates of **GNWT** and **IIT**, (Reardon, 2019) and (Melloni, Mudrik, Pitts, & Koch, 2021), the **CTM** shares features of both basic theories, as pointed out above. Our view is that both theories add to the discussion of consciousness. The adversarial aspects between the theories arise mainly from the advocates' differing views on brain regions primarily responsible for consciousness –prefrontal cortex for **GNWT**, posterior cortex for **IIT**. We note however, it is possible to have some **level** of consciousness without a cerebral cortex at all (Merker, 2007) and suspect that in such cases, as in the **CTM**, aspects of the basic **GWT** and **IIT** are still in play.

Our view on **free will** (section 3.5) is close to Dehaene's (Dehaene S. , Consciousness and the Brain: Deciphering How the Brain Codes Our Thoughts, 2014). Our explanation of the *feeling* of free will in the **CTM** incorporates additionally and *especially,* resource limits imposed by computational complexity considerations.

# Acknowledgements

We are especially grateful to Jean-Louis Villecroze for his comments, suggestions, and painstaking multiple reviews of our drafts, his pointers to the literature, and his ongoing work to simulate **CTM** (Villecroze, Personal Communication), (Villecroze, 2019). We thank Paul Liang for his insight and work with us, for teaching us about multi-modal machine learning, and for developing our multi-modal language Brainish (Liang, 2022). We thank the students and faculty at CMU and PKU for their feedback in our courses. We are grateful to our friend Michael Xuan for his enormous personal support and encouragement. We thank UniDT for their grant supporting our work. We also thank the reviewers of early versions of this paper who provided truly thoughtful constructive suggestions.

[52] In (Koch, The Feeling of Life Itself: Why Consciousness Is Widespread but Can't Be Computed, 2019), Christof Koch outlines the axioms: "[E]very conscious experience has five distinct and undeniable properties: each one exists for itself, is structured, informative, integrated and definite".

## FAQ

**Q1. Since a Universal Turing Machine (UTM) can simulate the CTM, won't it also be conscious?**

**A1. Short Answer.** The **UTM** can simulate the **CTM** as a Turing Machine, but not as a **Conscious Turing Machine**.

**Long Answer.** We argue that the **CTM** has the *feeling* it is conscious, not that it is conscious.
**CTM**'s "feeling of consciousness" does not come from its input-output function but from its architecture and the events happening inside the machine: its internal multi-modal language (**Brainish**), its predictive dynamics (cycles of prediction, feedback and learning), certain special processors (especially, **Model-of-the-World** and **Inner generalized Speech**), and from resource considerations.[53] Simulating the **CTM** on a universal Turing machine significantly alters those properties.

For example, conscious awareness in the **CTM** is defined as the *simultaneous* reception by all processors of a broadcast from the Short Term Memory (**STM**) stage and the subsequent immediate reinforcing ignition. The **UTM** simulation loses the simultaneity completely, and simultaneity is an essential component of the *feeling* of consciousness. The **UTM** will record the existence of a simultaneous reception but not experience it.

As with Tononi's Integrated information Theory (**IIT**) of consciousness (Tononi, 2004), completely unwinding the **CTM** would eliminate its consciousness (Kleiner, 2020).

---

**Q2. Since the CTM has Free Will, won't a UTM simulating the CTM also have Free Will?**

**A2.** We don't argue that the **CTM** has **Free Will** but that it has the ***feeling* of free will**. Recall (section 3.7):

> We define **free will** as the ability to make a decision that violates the laws of physics. From this it follows that no one has free will.
>
> We define the ***feeling* of free will** as the *conscious knowledge* that, when faced with a choice, the **CTM** (knows that it) can do computations to (help) make the decision, i.e., based on those computations and its utility (the game it is playing), it (knows that it) can make a better decision than can be made if it did not do the computations. That ***feeling* of free will** is something one can have.

As argued in **A1** above, a **UTM** simulating a **CTM** will not be a **CTM** as It has no *conscious knowledge* – no broadcasting of knowledge to all processors - and thus (from the definition of the feeling of free will) the **UTM** does not have the *feeling* of free will.

Without such a distinction between **Free Will** and the ***feeling* of free will**, the concept of **Free Will** is paradoxical. It was a puzzle at least as far back as Lucretius (100-50 BCE). And it led Samuel Johnson (1709-1784) to wonder how it can be that "All science is against the freedom of the will; all experience is for it."

---

**Q3. Is the CTM a blueprint for a conscious AI?**

**A3.** We view the **CTM** as a framework for understanding (animal or machine) consciousness, not a blueprint for designing a conscious AI. The **CTM** is a purposely simple stripped down substrate independent model for understanding consciousness. Indeed, to keep the **CTM** simple and understandable, adding features to it should be done only with the utmost care, and only if doing so is (arguably) necessary. That said, we do not claim that the **CTM** is the only possible machine model of consciousness, just as the Turing machine is not the only possible model of computation.

[53] Taking limited resources into account, in particular time limitations, is a theoretical computer science (**TCS**) feature that we have incorporated into the global workspace theory (**GWT**) in constructing the **CTM**.

**Q4. What is the advantage of having the CTM architecture?**

**A4.** When a processor doesn't know where to find the information it is looking for, a global broadcast requesting that information can get other processors engaged in finding it. That's one advantage. Multi-modal integration of information is another functionality achieved by this architecture.

---

**Q5. Why does the STM hold only one chunk?**

**A.** George Miller suggested the magic number 7±2 for the number of chunks in human short-term memory (Miller, 1956). Having a small number such as this is important for focusing attention. Some folks are incredulous that so few chunks will suffice. The **CTM** emphasizes this point by having just 1 chunk.

---

**Q6. Why is the CTM defined in the specific format given (and not some other)?**

**A.** Other formats may be just as good. Some format had to be chosen. Turing chose quintuples for his machines. Post used quadruples.

---

**Q7. What distinguishes conscious (STM) and unconscious (LTM) processes?**

**A**. Conscious processing is what gets broadcast from **STM** plus whatever communications about that broadcast through links between processors continue to keep it alive, what (Dehaene & Changeux, 2005) call "ignition". Unconscious processing is all the rest. Processing that does not reach **STM** does not get broadcast and therefore remains unconscious.

---

**Q8. Why is the Up-Tree a strictly binary tree?**

**A**. The binary **Up-Tree** could more generally be a **k**-ary tree for some small **k**, **k** much less than **N** (the number of processors). The original Turing Machine (**TM**) had just one read/write head and one 1-dimensional tape. Since then, others have considered **TM**'s with multiple tapes, multiple heads per tape, multi-dimensional tapes, and so on. In a similar way, the **Up-Tree** is made binary because binary is both *simple* and *sufficient*, and because the choice between 2 chunks at a node is slightly simpler to describe.

---

**Q9. Why is the resolution mechanism (in the Up-Tree competition) the specific one that is proposed.**

**A.** For the probabilistic **CTM**, the decision made at each interior node of the **Up-Tree** – namely which one of the node's two children's chunks should win the match - is decided by which chunk has the larger **f**-value. As processor **p**'s chunk works its way up the tree, its **f**-value is affected by those processors that neighbor **p**. It is a surprising consequence of this mechanism that if **f** is additive then the probability that a chunk rises to **STM** is independent of its location, that is to say, the location of the processor that generated it, on (the leaves of) the competition tree. As a consequence, the competition is permutation independent.

There are other completely different and even more important reasons for making **f** additive: for one, when **f** is not additive, in each and every one of the many examples considered, something always goes terribly wrong. For example, if **f (chunk) = |mood|** then a strong positive mood and a strong negative mood appearing in two siblings, children of a node, completely cancel: neither becomes conscious (i.e., reaches **STM**) even when all other chunks are relatively unimportant.

For example, consider the following **Up-Tree** with **f = |mood|** and $w_1 = 100, w_2 = -100, w_3 = 1, w_4 = 2$. Then the **Up-Tree competition** looks like this:

```
f = |mood|      Mood =                6  <—  Pr (A) = pr(B) = 0   Pr(C) = 1/3   Pr(D) = 2/3
                                    /    \
                 Mood =          0          6  <—  Pr(C) = 1/3   Pr(D) = 2/3
                               /   \       /  \
          Weight = Mood =    100  -100     2    4
                              A     B      C    D
```

For another example, if **f(chunk) = |weight|** (see figure in answer **A1** to **Q12**), then two chunks having the same maximum **|weight|** can have vastly different probabilities of reaching **STM**. This does not happen if **f** is additive.

---

**Q10. Why do you focus on the probabilistic rather than the deterministic CTM as being the correct model?**

**A.** There are *many* reasons for this. For one, as noted above, with any additive competition function such as **f(chunk) = intensity**, the competition is permutation independent if **CTM** is probabilistic. This is not the case if **CTM** is deterministic, not even if **f** is additive:

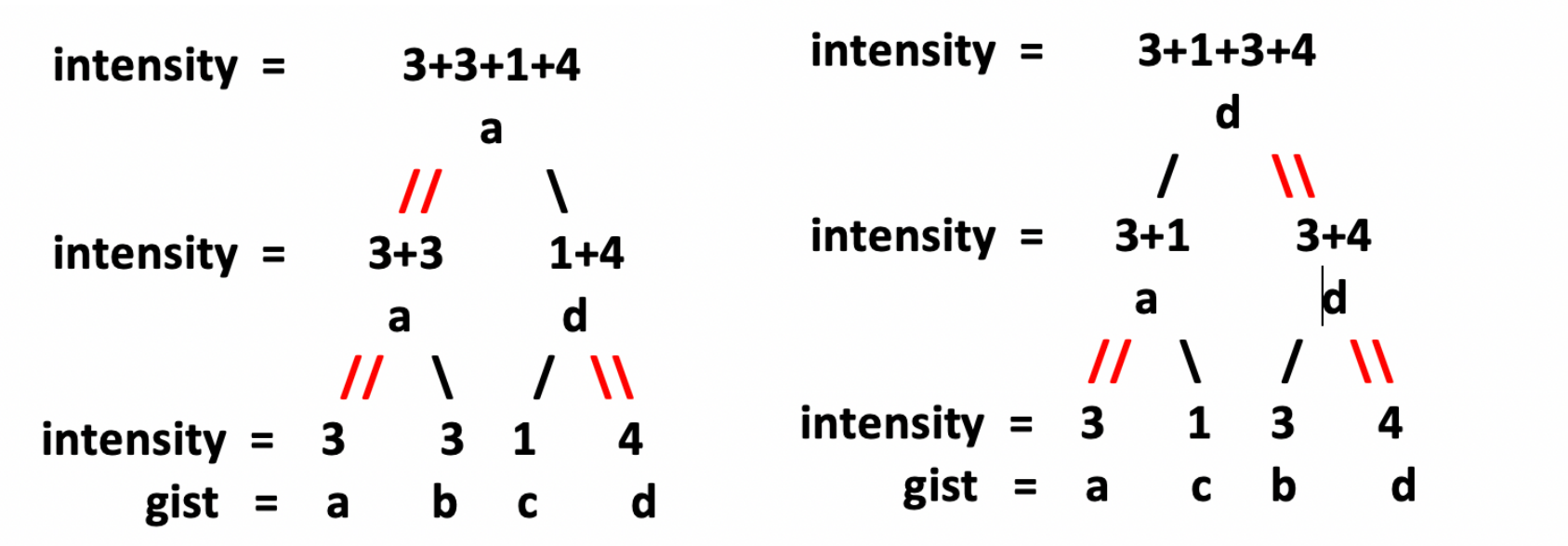

*Deterministic Competition with Competition Function **f: chunk → intensity**.*

For another, in the probabilistic **CTM**, gists submitted to the competition, even those with small intensity, get into **STM** with probability proportional to their estimated importance (**f**-value). Again, this is not the case for a deterministic **CTM**.

---

**Q11. Why an Up-Tree at all?**

**A.** We need a vehicle for **LTM** processors to get **CTM** to pay attention to the "most important" information. We chose the **Up-Tree** to execute the competition in part because it computes locally (between 2 siblings) to get the globally most important information into **STM.** With an additive competition function, the **Up-Tree** structure and competition ensure that processors get their chunks into **STM** with probability proportional to their **f**-value, which is arguably the correct way to do it. We consider the **Up-Tree** and the decisions it makes to be of fundamental importance.

---

**Q12. Why do the authors choose to have intensity and mood in the chunk? It seems equally valid to discard them.**

**A1**. Without **intensity** and **mood** in the chunk, every "reasonable" competition function such as **f(chunk) = |weight|** is non-additive, which leads to the possibility of a weird lopsided kind of consciousness. For example, suppose the competition tree has **N/2** chunks each of a *heavy* **weight W** in the left-hand subtree (**LHST**), exactly

**1** chunk of the same weight **W** in the right-hand subtree (**RHST**), and that all other chunks have *negligible* **weight**, as in the next figure. In that case, though **CTM** considers all heavily weighted chunks equally important, the (heavily weighted) chunks in the **LHST** each have negligible probability **1/N** to get into **STM**, while the single heavily weighted chunk in the **RHST** has probability almost **1/2** to get into **STM**:

```
weight =                 100        <------ pr{A} = pr{B} = pr{C} = pr{D} ≈ 1/8;   pr{E} ≈ 1/2
                     /          \
weight =         100            100
               /     \         /     \
weight =     100      100    100      1
            /  \     /  \   /  \    /  \
weight =   100 100 100 100  100  1  1    1
            A   B   C   D   E    F  G    H      <------ N=8
```

**A2.** At time **t+h**, the winning chunk contains the **weight** that was originally assigned to it at time **t** when it got put into the competition. The **intensity** and **mood** of that winning chunk, which were set to **|weight|** and **weight** respectively at time **t**, got continuously modified as the chunk moved up the competition tree until the chunk entered the **STM** at time **t+h**, at which time the **intensity** and **mood** are indicators of (**N** times) the average intensity and mood of the entire **CTM** at time **t**. We believe that humans are normally consciously aware of their global intensity and mood, a fact that makes it entirely reasonable to include **intensity** and **mood** in the chunk.

**A3.** Chess and tennis tournaments use seeding to give players of equal strength roughly equal chances of winning the tournament. The **Up-Tree competition** with *additive* competition function assures that even without seeding, all players have a probability of winning proportional to their ability/expertise.

---

**Q13. Where does feedback come from?**

**A.** Feedback comes from chunks that are received in **broadcasts** from **STM**, through **links**, and from the environment via **Input maps**, all of which have information that can be graded as “erroneous” or “correct” in comparison to (stored) predictions.

---

**Q14. How do processors judge whether or not their information is valuable?**

**A.** Judgements are based on feedback. Each processor has a **Sleeping Experts Algorithm (SEA)** that learns, based on feedback, what the weight-giving power of its processor should have been, so that weight assignments eventually settle down to something more or less correct. Roughly speaking, when a **|weight|** is too low, the **SEA** multiplies the weight-giving power by **2** (more generally by some constant **c > 1**). When too high, the **SEA** multiplies it by ½ (more generally by **1/c**).

---

**Q15. Why would chunks contain queries and answers?**

**A.** For example, when you meet a person at a party and can’t remember her name, a chunk produced by a processor can pose the query “What’s her name?”. When this chunk wins the competition for **STM**, its query is broadcast to all **LTM** processors. Sometime later, another processor answers, “I think her name begins with **T**,” which rises to **STM** and gets broadcast. This can later, perhaps much later, trigger another processor to answer, “Her name is **Tina**.”

---

**Q16: Must the CTM have a Model-of-the-World processor?**

**A1:** The **Model-of-the-World processor** is a fundamental component of consciousness. Our explanation for the "feeling of consciousness" is for a **CTM** that has a **Model-of-the-World**. We don't see how to do without it.

**A2.** Our argument that **CTM** *feels* conscious depends on it having a **Model-of-the-World processor** and is akin to the argument given by the Attention Schema of (Graziano, Guterstam, Bio, & Wilterson, 2020). We argue that **CTM's** *feeling* of consciousness, in the sense that the term is normally understood, is a consequence of the fact that what the **CTM** consciously knows of the world and of itself in the world *is* the **Model-of-the-World processor's** view of the world, and its view of the "**CTM**" in its **model of the world**; and that this view, which includes that the "**CTM**" is conscious, is broadcast to all processors.

---

**Q17. Isn't naming specific processors, such as the Model-of-the-World processor, in definite and final terms, an over-specification of the model?**

**A.** The proposed **Model-of-the-World processor** is only an example of how such a processor might work. It is not meant to be the definitive final specification. It's kind of like Turing's universal machine. The idea is important, as is having a description of such a machine. The particular machine is less important.

---

**Q18. Why would the whole cortex not constitute a Model-of-the-World processor (as commonly assumed in neuroscience) and not just the Model-of-the-World processor?**

**A.** The whole cortex may well be viewed as a model of the world in both the human brain and the **CTM**. We don't suggest otherwise.

---

**Q19. Why would the lack of input from the environment lead to incoherent thoughts in dreams?**

**A.** No, no, we're not saying that dreams *must* be incoherent, only that they *can* be incoherent, and that this is especially the case in dreams because the processors involved in a dream are not getting feedback from the environment. For example, in a dream, one might believe that one can fly.

---

**Q20. What question does the theory address that is not already accounted for by the standard GWT-related theories?**

**A1.** Unlike standard **GWT**-related theories of consciousness, **CTM** is a substrate independent computational model of consciousness, not a model of the brain. Its purpose is to explain how a machine can *experience* feelings. As Arlindo Oliveira has pointed out: "The proposed model is not a model of human consciousness, but a computational model that can explain many features of conscious behavior and that address directly the hard problem of consciousness, as defined by Chalmers. [It explains] why systems that are subject to the laws of physics can have subjective experiences."

**A2.** No other **GWT**-related theory gives a substantive idea how processors might decide among themselves what information to send to the stage.

---

**Q21. How does the theory argue that CTM has free will?**

**A.** We don't. We argue that **CTM** has the *feeling* of free will. Our argument is two-fold.

The first part of the argument has to do with resource limitations - a complexity theory argument. For example, when the **CTM** plays chess, it can be faced with a selection of possible moves but, not yet having evaluated the consequences of those moves, the **CTM** is free (and knows it is free) to choose whichever move it reckons best - within the time constraints.

The second part of the argument is that **CTM**'s **Model-of-the-World processor** tags the "**CTM**" in its **models-of-the-world** with a multi-modal Brainish gist asserting that **CTM** is in the process of choosing its next move - meaning that it (the **CTM**) is free to choose its next move. Of course, its decision is deterministic (assuming as we do Newton's deterministic physics). However, this labeling of **CTM** as having **free will** gives **CTM** its knowledge that its processors may now suggest the next move, and this knowledge is conveyed by a *feeling of* **free will**. This argument is similar to our argument that **CTM** *feels* it is conscious.

---

**Q22. Do you have an implementation of the CTM?**

**A**. We do not. That said, Jean-Louis Villecroze is working on an implementation as we speak, and Paul Liang is working on developing the multi-modal language, Brainish, in his PhD research on multi-modal machine learning. Our own focus is on understanding the hard problem of consciousness and some essential related features. We are not attempting to provide novel biological predictions, nor AI implementations - as wonderful as those would be - but to provide a simple machine model for consciousness.

We note that it took Alan Turing almost a decade (mostly due to his war work) to go from his theoretical one tape universal machine to his complete circuit specification for the implementation of a universal computer – a description so complete that it included vacuum tube choices, resistor and capacitor values, mercury delay line memories, and even the cost of the computer in pounds. Unfortunately, due to politics, Turing's Automatic Computing Engine (ACE) never saw the light of day; only a more primitive computer, the Pilot ACE, was constructed.

---

**Q23. Why is there little mention of neural correlates of consciousness, particularly with respect to the phenomenal aspects of consciousness?**

**A1.** The **CTM** is a computational substrate-independent model of consciousness, not a model of human or animal consciousness. (That said, a number of explanations from **CTM** of phenomenal aspects of consciousness such as blindsight are corroborated at a high level by cognitive neuroscience literature.)

**A2**. Even a complete knowledge of the "circuitry" of the brain and a complete knowledge of the neural correlates of consciousness – as wonderful and desirable as it would be to have these - cannot explain how the *feeling* of consciousness arises. To understand that feeling, something else is needed. That something is what we are proposing to get a handle on with the **CTM**.

---

**Q24. Could brain dynamics and competition between attractors be a more neurally plausible explanation for how the brain works?**

**A**. Perhaps, but... we're not looking to model the brain or brain dynamics but to understand consciousness. For this purpose, we use the mathematics that we find most helpful.

---

**Q25. Assuming the brain is a CTM, what are some conditions for a part of the brain to be considered the STM?**

**A.** The **STM** has a very small memory, a relatively small direct input, and an output that goes almost everywhere. In "A Brain Structure Looking for a Function" (Koch, 2014), Christof Koch suggests that the claustrum might fit the bill.

---

## **About the Authors** of the expanded monograph **(Blum, Blum, & Blum, monograph in preparation)**.

**Manuel** has been motivated to understand the mind/body problem since he was in second grade when his teacher told his mom she should not expect him to get past high school. As an undergrad at MIT, he spent a year studying Freud and then apprenticed himself to the great anti-Freud[54] neurophysiologist, Dr. Warren S. McCulloch, who became his intellectual mentor. When he told Warren (McCulloch) and Walter (Pitts) that he wanted to study consciousness, he was told in no uncertain terms that he was verboten to do so - and why (there was no fMRI at the time). As a graduate student, he asked and got Marvin Minsky to be his thesis advisor. Manuel is one of the founders of complexity theory, a Turing Award winner, and has mentored many in the field who have chartered new directions ranging from computational learning, cryptography, zero knowledge, interactive proofs, proof checkers, and human computation. He is a Fellow of $AAAS_1$, $AAAS_2$, NAS, NAE.
**Manuel Blum** mblum@cs.cmu.edu

**Lenore** has been passionate about mathematics since she was 10. She attributes that to having dropped out of school when she was 9 to wander the world, then hit the ground running when she returned and became fascinated with the Euclidean Algorithm. Her interests turned to non-standard models of mathematics, and of computation. As a graduate student at MIT, she showed how to use saturated model theory to get new results in differential algebra. Later, with Mike Shub and Steve Smale, she developed a foundational theory for computing and complexity over continuous domains such as the real or complex numbers. The theory generalizes the Turing-based theory (for discrete domains) and has been fundamental for computational mathematics.

Lenore is internationally known for her work in increasing the participation of girls and women in STEM and is proud that CMU has gender parity in its undergraduate CS program. Over the years, she has been active in the mathematics community: as President of the Association for Women in Mathematics, Vice-President of the American Mathematical Society, Chair of the Mathematics Section of the American Association for the Advancement of Science, Deputy Director of the Mathematical Sciences Research Institute, and as Inaugural and current President of the Association for Mathematical Consciousness Science (AMCS). She is a Fellow of AAAS, AMS, AWM.
**Lenore Blum** lblum@cs.cmu.edu

**Avrim** had an earlier start than the elder Blums. He spent his first two years at MIT, in his mom's office in the Math Department, and in his dad's office in McCulloch's lab. In sixth grade, he solved an extra credit math problem by programming his home-made computer to get a feel for the problem, then (once he saw what was going on) stated and proved the desired result. Because he used a computer, he got no credit. Odd, because he was pointing to a novel way (at the time) to solve a math problem. Avrim's expertise is Machine Learning Theory. He has been an advisor to many of the young leaders in the field.

Avrim is an active member of the computer science community. He has served as Program Chair for the IEEE Symposium on Foundations of Computer Science (FOCS), the Innovations in Theoretical Computer Science Conference (ITCS), and the Conference on Learning Theory (COLT). He has served as Chair of the ACM SIGACT Committee for the Advancement of Theoretical Computer Science and on the SIGACT Executive Committee. He is recipient of the AI Journal Classic Paper Award, the ICML/COLT 10-Year Best Paper Award, a Sloan Fellowship, the NSF National Young Investigator Award, and the Herbert Simon Teaching Award. He is a Fellow of the ACM.
**Avrim Blum** avrim.blum@gmail.com

All three Blums received their PhDs at MIT and spent a cumulative 65 wonderful years on the faculty of the Computer Science Department at CMU. Currently the elder two are emeriti and the younger is Professor and Chief Academic Officer at TTIC (Toyota Technological Institute at Chicago), a PhD-granting computer science research institute focusing on areas of machine learning, algorithms, AI (robotics, natural language, speech, and vision), data science and computational biology, and located on the University of Chicago campus. Manuel Blum is Emeritus Professor of Computer Science at UC Berkeley and CMU. Lenore Blum has been a Distinguished Career Professor of Computer Science at CMU and is currently a Distinguished Professor-in-Residence at UC Berkeley.

[54] Where Freud had written *The Future of an Illusion* (Freud S. , 1927) , McCulloch followed with "The Past of a Delusion" (McCulloch W. S., 1953).